\documentclass{article}
\usepackage{spconf,amsmath,graphicx}
\usepackage{amssymb}
\usepackage{array}
\usepackage[caption=false,font=normalsize,labelfont=sf,textfont=sf]{subfig}
\usepackage{textcomp}
\usepackage{stfloats}
\usepackage{url}
\usepackage{verbatim}
\usepackage{graphicx}
\usepackage{algorithm}
\usepackage{algpseudocode}
 \usepackage{booktabs}
\usepackage{colortbl}
\usepackage{xcolor}
\usepackage{balance}
\definecolor{lightergray}{gray}{0.95}
\usepackage{multirow}
\usepackage{siunitx}
\title{Cross-Chirality Palmprint Verification: Left is Right for the Right Palmprint}

\name{Chengrui Gao$^{\dagger}$, Ziyuan Yang$^{\dagger}$, Tiong-Sik Ng$^{\ddagger}$, Min Zhu$^{\dagger \star}$ \thanks{${\star}$: Corresponding author}, Andrew Beng Jin Teoh$^{\ddagger \star}$}
  
\address{$^{\dagger}$ College of Computer Science, Sichuan University, Chengdu 610045, China \\
      $^{\ddagger}$ School of Electrical and Electronic Engineering, Yonsei University, Seoul 120749, South Korea}

\begin{document}
%

\maketitle

%
\begin{abstract}

Palmprint recognition has emerged as a prominent biometric authentication method, owing to its high discriminative power and user-friendly nature. This paper introduces a novel Cross-Chirality Palmprint Verification (CCPV) framework that challenges the conventional wisdom in traditional palmprint verification systems. Unlike existing methods that typically require storing both left and right palmprints, our approach enables verification using either palm while storing only one palmprint template. The core of our CCPV framework lies in a carefully designed matching rule. This rule involves flipping both the gallery and query palmprints and calculating the average distance between each pair as the final matching distance. This approach effectively reduces matching variance and enhances overall system robustness. We introduce a novel cross-chirality loss function to construct a discriminative and robust cross-chirality feature space. This loss enforces representation consistency across four palmprint variants: left, right, flipped left, and flipped right. The resulting compact feature space, coupled with the model's enhanced discriminative representation capability, ensures robust performance across various scenarios. We conducted extensive experiments to validate the efficacy of our proposed method. The evaluation encompassed multiple public datasets and considered both closed-set and open-set settings. The results demonstrate the CCPV framework's effectiveness and highlight its potential for real-world applications in palmprint authentication systems.

\end{abstract}

\section{Introduction}
Palmprint recognition has gained significant traction as a robust biometric modality in recent years, demonstrating its efficacy in diverse real-world applications~\cite{su2023learning,fan2023amgnet}. 

In traditional palmprint recognition systems, users usually register their left and right palmprints~\cite{jing2021double}, as illustrated in Fig.~1(a). This dual enrollment provides the system with additional data in identifying or verifying individuals, as matching between palms is not feasible. This approach is particularly valuable because it ensures reliable recognition even if one hand's data is unavailable or compromised. Furthermore, since some individuals may have more distinctive features in one palm than the other, capturing both improves recognition accuracy, reduces error rates, and boosts security. It is especially effective for deployments where reliability is paramount.

However, this design also brings certain drawbacks. Storing both palmprints doubles the storage requirements, which can be problematic for systems with extensive databases. Processing two sets of biometric data adds computational overhead, potentially leading to longer response times during enrollment and matching. Additionally, it can be inconvenient for users who must provide prints from both palms, potentially impacting usability. Finally, storing more biometric data for each user increases the potential damage in the case of a data breach. If both palmprints are compromised, it can create a larger security issue than systems that store only one biometric modality.

\begin{figure}
  \centering
  \begin{minipage}[t]{0.9\linewidth}
    \centering
    \footnotesize
    \includegraphics[width=\textwidth]{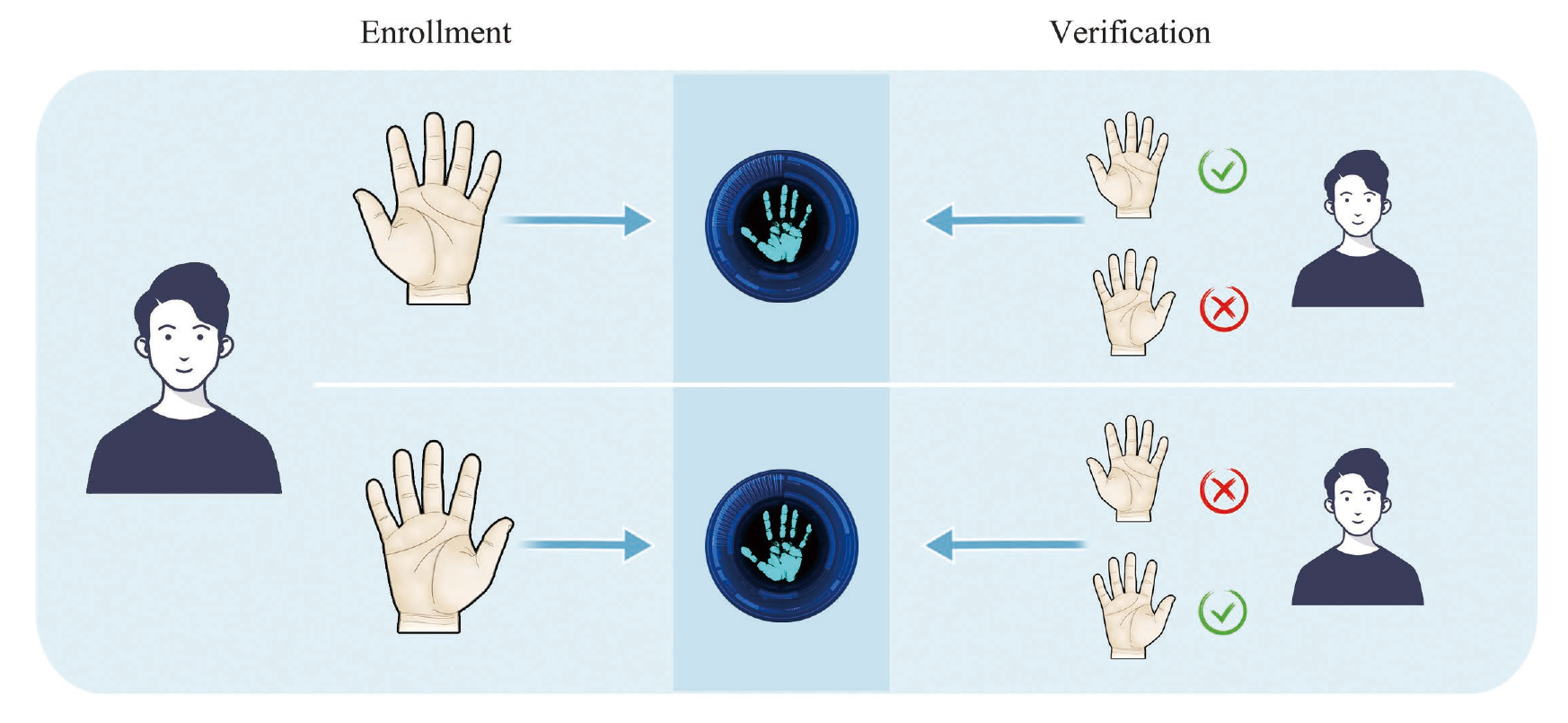}
    \centerline{(a) Traditional Palmprint Verification System}
  \end{minipage}  
  \begin{minipage}[t]{0.9\linewidth}
    \centering
    \footnotesize
    \includegraphics[width=\textwidth]{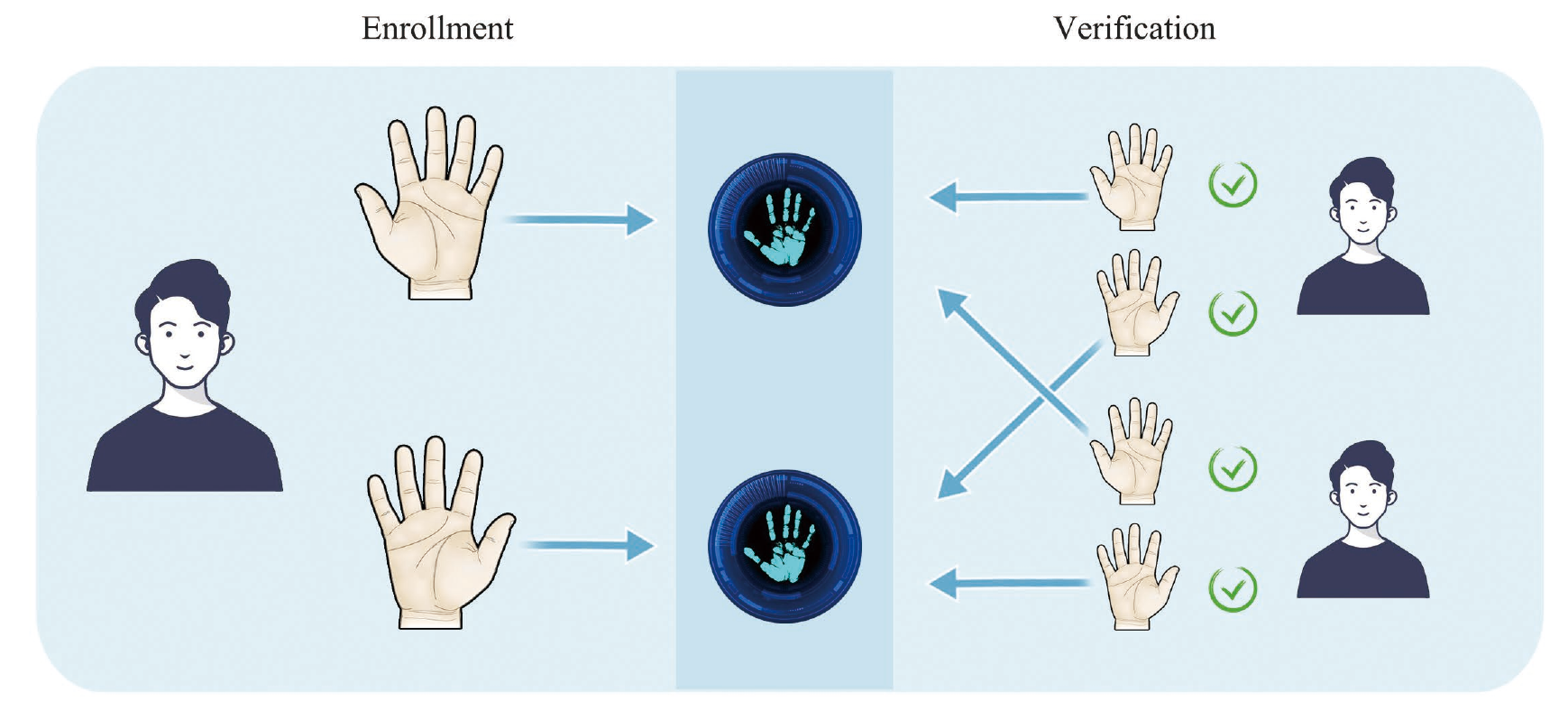}
    \centerline{(b) Our Proposed Cross-Chirality Verification System}
    \label{fig:our_frame}
  \end{minipage} 
  \caption{(a) Traditional Palmprint Verification System. (b) Our Proposed Cross-Chirality Verification System.}
  \vspace{-5pt}
\end{figure}

In this paper, we introduce a cross-chirality palmprint verification (CCPV) framework that stores only one palmprint-either left or right, allowing verification with either palm, as depicted in Fig. 1(b). This notion is rooted in biological principles, where genetic and molecular signals, such as those from Hox genes, regulate the symmetrical development of structures during embryonic growth. These genes are consistently expressed on both sides of the body, contributing to similar patterns in left and right palmprints~\cite{cumplido2024hox}. However, subtle developmental variations and environmental factors result in slight asymmetries between the two palms, making cross-chirality palmprint verification challenging despite their inherent structural similarities.

CCPV brings several key advantages. By storing only one palmprint, the system reduces storage needs and computational complexity, effectively halving the biometric data it must manage. This optimization improves response times and enhances system efficiency, particularly in large-scale deployments. Additionally, the flexibility of allowing users to verify with either palm increases convenience, making the system more adaptable to real-world scenarios where one hand may be unavailable or impaired. By simplifying the enrollment process to capture only one palmprint, user effort is minimized, boosting system usability and adoption rates. Furthermore, focusing on a single palmprint for matching reduces the risk of errors or mismatches due to environmental factors, as the system doesn’t need to reconcile two sets of biometric data.

To implement CCPV, one might consider assigning the same identity label to an individual's left and right palmprints and training the palmprint verification model. However, this approach faces significant challenges, such as model collapse, due to the inherent differences between left and right palmprints. Moreover, it fails to drive the model to learn cross-chirality features effectively, as it lacks constraints in feature space construction. This leads to suboptimal performance in open-set scenarios—a scenario where training and deployment datasets are disjoint, and generalization is crucial.

To address the cross-chirality matching problem challenge, we reconsider the traditional approach that directly compares the dissimilarity between a probe and a query instance. Drawing inspiration from biological insights on the symmetrical nature of palmprints between the left and right hands, we propose an extension to the matching process by moving from a single-matching rule to a four-matching rule. Specifically, we introduce a method that flips both the probe and query instances, which helps minimize structural differences and leverages the natural symmetry of palmprints. This results in two flipped queries and two flipped probes, yielding four potential matches. 

This approach is driven by two key considerations. First, since we cannot always know whether the probe and query belong to the same hand, it is crucial to account for both the similarity between reversed palm structures and the possibility that they originate from the same hand. Second, incorporating this multi-match strategy during training acts as a form of data augmentation, helping to prevent overfitting. By conducting multiple matches, we reduce variance and more effectively capture the detailed texture features of the palmprints.

We introduce a Chirality-Consistency Loss (CC Loss) to simulate the matching inference process during training. Specifically, we sample each individual's left and right palmprints, annotate them with the same label, and treat them as a single user. Each training batch is then composed of multiple individual samples. Following the inference rules, we aim to construct a compact feature space for genuine matches and more dispersed for imposters. Our constraint is based on the four matching pairs, allowing us to build a robust, cross-chirality feature space that ensures the network learns to extract symmetrical features from palmprints. Rather than focusing on direct classification, emphasizing representation learning at the feature level enhances the model's open-set recognition capabilities.

Our main contributions can be summarized as follows:

\begin{itemize}

\item We introduce a CCPV framework that stores only one palmprint, enhancing system efficiency, user convenience, and accuracy while reducing storage, computational complexity, and the risk of mismatches.

\item We propose a four-matching rule to address the structural similarity between a hand and its reversed counterpart, considering that both prints may originate from the same hand during training.

\item We present a novel CC Loss designed to create a compact feature space for genuine samples while keeping a more dispersed feature space for imposters, ensuring the network consistently extracts robust features across different hand orientations.

\item Extensive experiments are conducted on public palmprint datasets, and the results demonstrate that the proposed paradigm performs well with different backbones in the cross-chirality palmprint verification setting in both close-set and open-set scenarios.
\end{itemize}

\section{Related Works}

\subsection{Palmprint Recognition}

Palmprint recognition technology is widely popular in various applications due to its user-friendliness, privacy, and high discriminability~\cite{liu2022data,du2020cross}. The palmprint verification methods can be broadly classified into four categories: subspace-based methods, statistical-based methods, coding-based methods, and deep learning-based methods~\cite{zhong2019decade}. Subspace-based methods aim to design projection formulas that map palmprint images into a low-dimensional subspace~\cite{fei2020feature}. Statistical-based methods typically involve extracting features from palmprint images and using statistical techniques to extract discriminative information~\cite{zhang2018combining}. Coding-based methods are designed to extract distinctive texture features for verifying individual identity~\cite{yang2020extreme}.  For example, Zhang~\textit{et al.}~\cite{zhang2003online} proposed PalmCode, which uses a 2D Gabor phase encoding scheme to extract and represent palmprint features, achieving satisfactory performance. Inspired by this, numerous variants have been developed~\cite{fei2020learning}. Yang \textit{et al.}~\cite{yang2023multi} combined first-order and second-order feature extraction to achieve better performance.

However, these methods were designed based on prior knowledge, limiting their recognition performance and robustness. 
In recent years, feature representation based on deep learning has become mainstream in palmprint recognition, offering higher accuracy and robustness compared to traditional handcrafted descriptors~\cite{jin2024pce}. With the success of deep learning in various tasks~\cite{yang2023hypernetwork}, researchers have shown increasing enthusiasm for incorporating related technologies into palmprint recognition. For example, Liang~\textit{et al.}~\cite{liang2021compnet} proposed the trainable Gabor filters-based palmprint recognition network~(CompNet). Besides, Yang \textit{et al.}~\cite{yang2023comprehensive} proposed a comprehensive competition mechanism for deep features, considering the spatial and orientation competition features. Jia \textit{et al.}~\cite{jia2022eepnet} proposed a lightweight deep network (EEPNet) for palmprint recognition, embedding two additional losses into an improved MobileNet-V3. Zhao \textit{et al.}~\cite{ma2023multiscale} introduced the Multiscale Multidirectional Binary (MSMDB) pattern learning method, enhancing palmprint recognition accuracy by maximizing the variance of learned binary codes between classes and minimizing intra-class distances. These methods fully capture the characteristics of palmprints to improve the overall performance of palmprint recognition. However, these single-modal recognition methods may have limitations in handling environmental changes and interference.

\subsection{Multi-Instance Fusion Recognition}

Researchers have also investigated the recognition of different instances within the same modality compared to traditional unimodal biometric recognition. Multi-instance biometric recognition has attracted significant attention due to its enhanced stability in recognition.

Yang \textit{et al.}~\cite{yang2024efficient} propose a Finger Disentangled Representation Learning Framework that separates each finger modality into shared and private features, enhancing fusion and extracting modality-invariant features for heterogeneous recognition. Consequently, researchers have explored combining left and right palmprint recognition to improve accuracy. Xu \textit{et al.}~\cite{xu2014combining} proposed a framework integrating left and right palmprints and cross-matching them for identity verification. Building on this, Jenifer and Kavidha~\cite{jenifer2015combining} introduced a multi-biometric recognition framework that combines left and right palmprint images at the matching score level. 

Ahmed \textit{et al.}~\cite{ahmed2024security} further advanced this by fusing left and right palmprint geometry features with palmprint features to create fusion feature vectors. Their matching/normalization module utilized cosine similarity and corrected distance methods to match hand geometry and palmprint features. However, these methods require users to provide left and right-hand information during the deployment (enrollment and verification) phases.

In contrast to traditional multi-instance approaches, we propose a cross-chirality recognition method that does not necessitate the availability of both modalities during the deployment stage. Our method employs cross-chirality palmprint recognition, training the model containing both left and right palmprints to build an identity-level feature space. This allows for identity recognition by providing either a left or right palmprint during the verification when only the left or right palmprint is registered.

\subsection{Cross-Modal Recognition}

Several cross-modal biometric systems have been proposed among different biometric modalities such as palmprint, and palmvein~\cite{khatri2023deep,fan2024novel,rane2021dual}. For instance, Su \textit{et al.}~\cite{su2023learning} introduced a modality-invariant binary features learning method for cross-modality recognition between palmprint and palm-vein. This method projects images into a high-dimensional space to mitigate the effects of misalignment in heterogeneous data.

Dong \textit{et al.}~\cite{dong2022co} proposed PalmCohashNet, where each palmprint modality is collaboratively trained to generate shared hash codes for each modality. Additionally, a cross-modality hashing (CMH) loss is designed to minimize the modality gap between palmveins and palmprints.
Kumar~\cite{kumar2016identifying} \textit{et al.}~\cite{kumar2016identifying} employed CNN networks for feature extraction and matched left and right palmprint images using a novel reference point selection approach combined with an edge detector. Although this algorithm demonstrated the potential for matching left and right palmprints, its effectiveness is limited to closed-set settings. However, this algorithm's traditional manual feature matching has limitations. It is feasible only in closed-set conditions and yields suboptimal results.

Most existing cross-modal biometric recognition models highlight the differences between biometrics across various modalities. In contrast, this paper introduces a cross-instance verification approach specifically designed for cross-chirality palmprint recognition. 

We propose a novel method for cross-left-right palmprint recognition based on feature template consistency, leveraging the natural symmetry between the left and right hands. This approach enhances accuracy and boosts its practical application potential by minimizing intra-class distances among left-left, right-right, and left-right templates.

\section{Methology}


\begin{figure*}
  \centering
  \includegraphics[width=\textwidth]{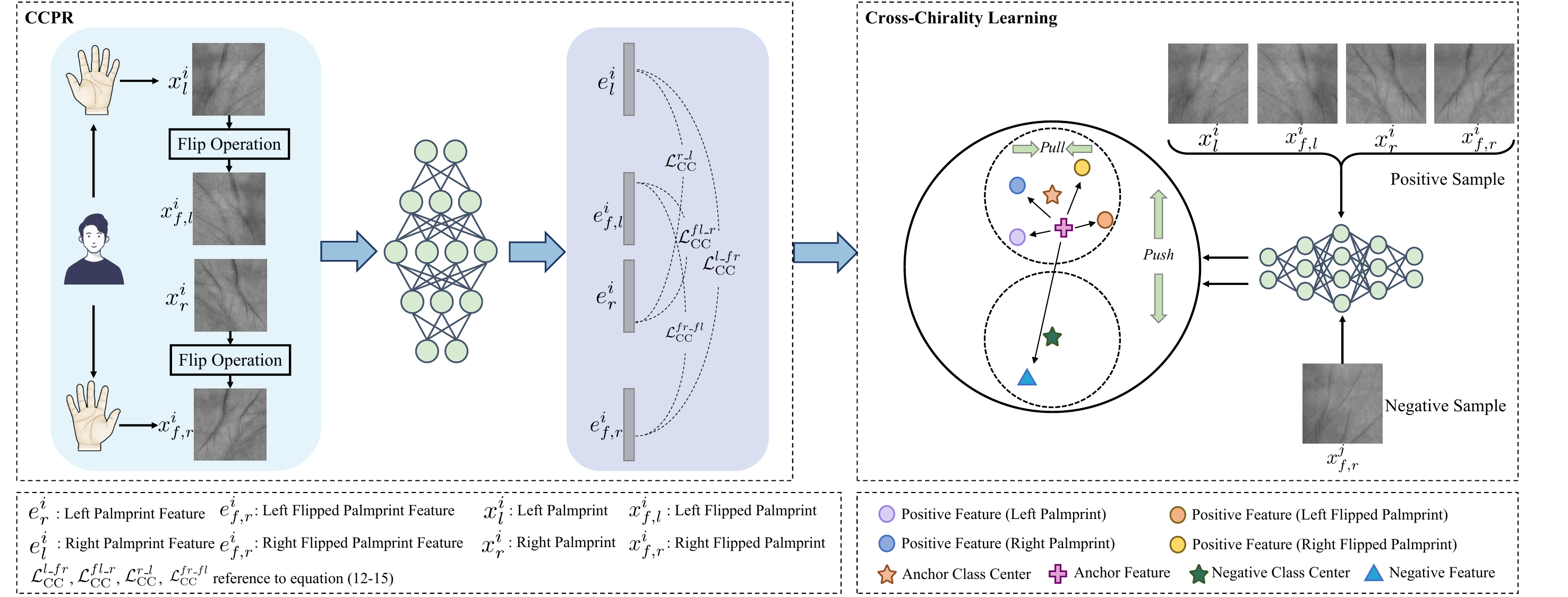}
     \caption{The whole framework of the proposed CCPV.}
  \label{fig:malicious}
\end{figure*}

\subsection{Problem Statement}

The traditional learning-based palmprint recognition problem~\cite{yang2023comprehensive,zhong2019centralized} can be formulated as follows:

\begin{equation}
\begin{aligned}
\arg\min_{\theta} \Big( 
& \text{dis}\left(f\left(x_{i,l}^g; \theta\right), f\left(x_{i,l}^q; \theta\right)\right) \\
& - \text{dis}\left(f\left(x_{i,l}^g; \theta\right), f\left(x_{j,l}^q; \theta\right)\right)
\Big)
\end{aligned},
\end{equation}
where $f$ denotes the recognition network, and $\theta$  represents its parameters. $x_{i,l}^g $ and $x_{i,l}^q $denotes the palmprint image from  $i$-th user in a gallery set and query set, $x_{i,l}^q $ denotes the query palmprint image from  $j$-th user, with $i \neq j$. $\text{dis}$ denotes the matching function.

In this paper, we propose a cross-chirality palmprint verification framework with a more stringent optimization objective, which can be formulated as follows:
\begin{equation}
        \arg\min_{\theta} (\text{dis}(f(x_{i}^g; \theta),f(x_{i}^q; \theta))),
\end{equation}
where $x_{i}^g$ denotes left-right palmprint from $i$-th user in gallery set, $x_{i}^q$ denotes left-right palmprint from $i$-th user in query set.

\subsection{Overview}
As previously discussed, traditional palmprint recognition focuses on distinguishing between different palmprints. In this paper, we propose a novel approach: cross-chirality palmprint recognition. While the human body is generally symmetrical, slight genetic differences and environmental factors during development lead to deviations between the left and right palmprints. To address this asymmetry, we simultaneously flip the query and probe images during the training stage, generating two query templates and two probe templates. These templates are then matched, producing four matching outcomes, with the final result calculated as the average of the matching distances.

We design a training framework that simulates the inference phase and introduces a CC Loss to build a discriminative feature space capable of handling cross-chirality matching. Our training framework and inference overview can be found in Fig.~\ref{fig:malicious}.

\subsection{Method}

A deep learning-based palmprint recognition framework typically comprises two key components: a recognition network and a loss function. In our CCPV framework, the matching rule and loss function take precedence over the network architecture itself. Therefore, this paper focuses primarily on these two critical aspects.

\begin{figure}
  \centering
  \includegraphics[width=0.6\columnwidth]{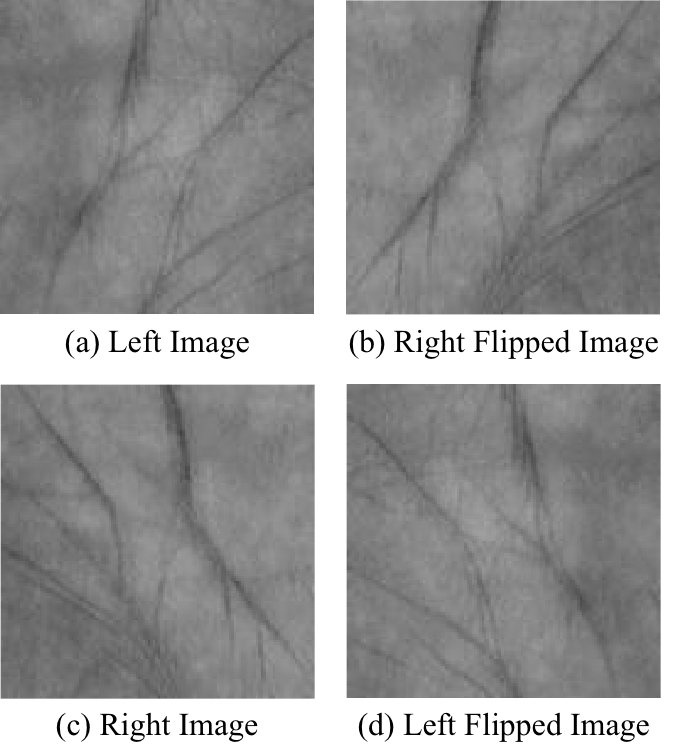}
     \caption{Sample palmprint images of the same individual from the Multi-Spectral dataset, with images flipped to highlight the similarity between palmprints from both hands.}
  \label{fig:fig3}
\end{figure}

Inspired by the symmetrical structures in human anatomy, we observe a similar pattern in palmprints, as shown in Fig.~\ref{fig:fig3}. Flipping a palmprint can produce an image that resembles the palmprint of the opposite hand. The main challenge in cross-chirality palmprint recognition is leveraging this symmetry effectively in the matching process. However, directly flipping the image is insufficient due to subtle yet significant differences between palmprints. Therefore, it is essential to design a novel loss function that enables the recognition network to construct a cross-chirality feature space. This loss function must also ensure strong discriminative power by compacting intra-class features while separating inter-class features.

\subsection{Four-Matching Rule}
 
 The four-matching rule is applied during the training phase to exploit the symmetrical nature of palmprints, helping to reduce structural inconsistencies and minimize variation in matching outcomes. This approach enhances recognition accuracy. Specifically, both the query and probe images are flipped, and the recognition network processes the resulting images to generate their respective templates. These query templates are then compared against the gallery templates, producing four distinct matching distances.

In practical scenarios, whether the query and probe palmprints come from the same hand is uncertain. Therefore, it is essential to account for both the structural similarity of reversed hand patterns and the chance that they may belong to the same hand. According to our matching rule, at least two of the four matching pairs will display structural similarity. In palmprint recognition, this structural alignment translates into textural similarity between the prints.

The remaining matching pairs serve as a form of regularization. In the worst-case scenario, the matching distances for these pairs might be quite large. However, since this regularization term is integrated into each authentication process, it does not significantly affect the overall recognition outcome. From a distributional standpoint, although intra-class and inter-class matching distributions may shift toward larger distances, their relative gap remains mostly unchanged. To further mitigate any potential negative impact from this regularization term, we have devised a specialized loss function, which will be detailed in the following subsection.

To eliminate ambiguity, we define $\dagger = \{l, r\}$, where $l$ and $r$ represent the left and right palmprints, respectively. Since it is uncertain which palmprint is used during enrollment or verification, we denote the gallery image as $x^g_\dagger$ and the query image as $x^q_\dagger$. In the following formulation, the subscript $\dagger$ indicates that the element is randomly chosen from $\dagger$, with a fresh selection being made for each variable when referenced.

First, the gallery and query images must be flipped to prepare for the subsequent matching step. This process can be expressed as follows:

\begin{equation}
\begin{aligned}
   x^g_{f,\dagger} &= \text{FLIP}(x^g_\dagger), 
\end{aligned}
\end{equation}
\begin{equation}
\begin{aligned}
   x^q_{f,\dagger} &= \text{FLIP}(x^q_\dagger),
\end{aligned}
\end{equation}
where $x^g_{f,\dagger}$ and $x^q_{f,\dagger}$ are flipped gallery and query images, respectively. $\text{FLIP}(\cdot)$ denotes the flip operation. As emphasized earlier, though $\dagger$ appears in $x^g_\dagger$ and $x^q_\dagger$, the sampled element could be different.

Next, we obtain the set $\mathcal{X} = \{x^g_{\dagger}, x^g_{f,\dagger}, x^q_{\dagger}, x^q_{f,\dagger}\}$, which includes the gallery and query images, both in their original and flipped forms. From this set, we construct the matching pairs: $(x^g_{\dagger}, x^q_{\dagger})$, $(x^g_{\dagger}, x^g_{f,\dagger})$, $(x^g_{f,\dagger}, x^q_{\dagger})$, and $(x^g_{f,\dagger}, x^q_{f,\dagger})$. 

Notably, regardless of which element is selected from $\dagger$ for the gallery and query images, at least two matching pairs within $\mathcal{X}$ will exhibit structural similarity.

The next step is to compute the distance for each matching pair. The final matching results are obtained as follows:

\begin{equation}
\begin{cases}
d_1 = \text{dis}(e^g_{\dagger},e^q_{\dagger}) \\
d_2 = \text{dis}(e^g_{\dagger},e^g_{f,\dagger}) \\
d_3 = \text{dis}(e^g_{f,\dagger},e^q_{\dagger}) \\
d_4 = \text{dis}(e^g_{f,\dagger},e^q_{f,\dagger}) \\
\end{cases},
\label{eq_dddd}
\end{equation}
where $d_1$, $d_2$, $d_3$, and $d_4$ denote the matching results of different matching pairs, respectively. $e^g_{\dagger},e^q_{\dagger}, e^g_{f,\dagger},e^q_{f,\dagger}$ represent the feature vector of $x^g_{\dagger}, x^q_{\dagger}, x^g_{f,\dagger}, x^q_{f,\dagger}$, respectively. 
$\text{dis}$ denotes the matching distance formulation. In this paper, $\text{dis}$ is defined as follows:
\begin{equation}
    \text{dis}(e_g,e_q) = \arccos\left(\frac{e_g\cdot e_q}{\|e_g\| \|e_q\|}\right) / \beta,
\end{equation}
where $e_g$ and $e_q$ denote the gallery and query features, respectively. $\beta$ denotes the temperature parameter.

Based on Eq.~\eqref{eq_dddd}, we could calculate the final matching result as follows:
\begin{equation}
    d = \frac{1}{4}\sum_{i=1}^{4} d_i,
\end{equation}
where $d$ is the final matching result of $x^g_{\dagger}$ and $x^q_{\dagger}$.

As previously discussed, the distances $d_1$, $d_2$, $d_3$, and $d_4$ can be divided into groups based on whether the corresponding images share structural similarity, namely ``similar group'' and ``opposite group''. In the absence of optimization constraints pertaining to model consistency, the matching results derived from the similar group yield more significant insights. In this case, assuming that the matching distances in the opposite group will be larger is reasonable. Since this assumption is consistently present in every matching process, the overall matching distributions remain unchanged, ensuring that verification performance remains unaffected.

\subsection{Loss Function}

As previously mentioned, the worst-case scenario occurs when no optimization term is applied to reduce the matching distance in the opposite group. To address this, we propose a novel loss function to enforce consistency in the templates, even when the images do not share a similar structure. By achieving this, the variance in matching distances can be minimized.

Building on previous work~\cite{yang2023comprehensive}, we employ cross-entropy loss $\mathcal{L}_{CE}$ to align samples belonging to the same class with their class center. To ensure template consistency, we introduce a CC loss. This ensures that the left palmprint image $x^i_l$, right palmprint image $x^i_r$, flipped left palmprint image $x^i_{f,l}$, and flipped right palmprint image $x^i_{f,r}$ for the $i$-th user share the same feature template. Achieving this makes the matching results in the opposite group meaningful.

Additionally, we use supervised contrastive loss~\cite{khosla2020supervised} to minimize the matching distances between pairs during training. This can be formulated as follows:

\begin{equation}
\mathcal{L}_{\text{CC}}^{l\_fr} = - \sum_{i \in I} \frac{1}{|P(i)|} \sum_{p \in P(i)} \log \frac{\exp(e_l^i \cdot e_{f,r}^p / \tau)}{\sum_{a \in A(i)} \exp(e_l^i \cdot e_l^a / \tau)},
\end{equation}
\begin{equation}
\mathcal{L}_{\text{CC}}^{{fl\_r}} = - \sum_{i \in I} \frac{1}{|P(i)|} \sum_{p \in P(i)} \log \frac{\exp(e_{f,l}^i \cdot e_r^p / \tau)}{\sum_{a \in A(i)} \exp(e_{f,l}^i \cdot e_{f,l}^a / \tau)},
\end{equation}
\begin{equation}
\mathcal{L}_{\text{CC}}^{r\_l} = - \sum_{i \in I} \frac{1}{|P(i)|} \sum_{p \in P(i)} \log \frac{\exp(e_r^i \cdot e_l^p / \tau)}{\sum_{a \in A(i)} \exp(e_r^i \cdot e_r^a / \tau)},
\end{equation}
\begin{equation}
\mathcal{L}_{\text{CC}}^{fr\_fl} = - \sum_{i \in I} \frac{1}{|P(i)|} \sum_{p \in P(i)} \log \frac{\exp(e_{f,r}^i \cdot e_{f,l}^p / \tau)}{\sum_{a \in A(i)} \exp(e_{f,r}^i \cdot e_{f,r}^a / \tau)},
\end{equation}
where $e_l^i, e_r^i, e_{f, l}^i, e_{f, r}^i$ be the feature of $x_l^i, x_r^i, x_{f, l}^i, x_{f, r}^i$, respectively. \( I \equiv \{1 \ldots 2n\} \) represents the batch of contrastive sample pairs, \( p \) denotes a sample that belongs to the same class as \( i \). \( A(i) \equiv I \setminus \{i\} \) is the set of all samples in the batch excluding \( i \) itself, with \( i \) being the index of the positive sample. \( P(i) \equiv \{p \in A(i): y^i = y^p\} \) is the index set of positive samples within the batch that are distinct from \( i \), where \( y^i \) is the label of the \( i \)-th sample in the batch. \( |P(i)| \) represents the number of samples in \( P(i) \). Lastly, \( \tau \) denotes the temperature parameter.

Then, the CC loss can be calculated as follows:

\begin{equation}
\begin{aligned}
    \mathcal{L}_{\text{CC}} = \frac{1}{4} \times \mathcal{L}_{\text{CC}}^{l\_{fr}}+\frac{1}{4} \times \mathcal{L}_{\text{CC}}^{{{fl}\_r}}+\frac{1}{4} \times \mathcal{L}_{\text{CC}}^{r\_l}+\frac{1}{4} \times \mathcal{L}_{\text{CC}}^{fr\_{fl}},
    \label{eq:ccloss}
\end{aligned}
\end{equation}
where $\mathcal{L}_{\text{CC}}$ denotes the CC loss.

By leveraging Eq.~\eqref{eq:ccloss}, we can achieve feature consistency across the four palmprint images of an individual. This ensures that each matching pair becomes meaningful, as the features of these images exhibit intra-class consistency while remaining distinguishable from inter-class comparisons. As a result, more matching pairs reduce variance and lead to more stable verification performance. The overall loss during the training phase can be formulated as follows:

\begin{equation}
    \mathcal{L} = w_{ce} \times \mathcal{L}_{CE} + w_{cc} \times \mathcal{L}_{CC},
\end{equation}
where $\mathcal{L}$ is the final loss, $w_{ce}$ and $w_{cc}$ denote the weights of $\mathcal{L}_{CE}$ and $\mathcal{L}_{CC}$, respectively.

The key steps are outlined in Alg.~\ref{alg} to detail our training process. During training, we sample the left and right palmprints of the same identity simultaneously. The palmprints are flipped, and the CC loss is applied to train the network. This approach enables the network to learn a consistent feature representation for the left and right palmprints, thereby reducing the matching distance in the opposite group. Importantly, although the left and right palmprints are sampled together during training, our method is not based on multi-instance recognition.

In the verification phase, we no longer require both palmprints simultaneously, as their consistency has been well-established during training. Our open-set experiments, detailed in Sec. IV.B further supports this, demonstrating that our method maintains strong performance even when tested on identities unseen during training. 

Additionally, the proposed CCPV framework is highly adaptable, with no specific requirements for the recognition network. It can be integrated with various recognition networks, as evidenced by the experiments discussed in Sec. IV.B.

\begin{algorithm}[!t]
\caption{Main Training Steps of the CCPV}
\label{alg}
\begin{algorithmic}[1]

\Function{MAIN}{}
\State Initialize model $f$
\For{$e = 1, 2, 3, \dots, T$} 
    \For{($x_l^i, x_r^i, y^i$) in $\mathcal{D}$}
        \State $x_{f,l}^i \gets \text{FLIP}(x_l^i)$ \Comment{Based on Eq. (6)} 
        \State $x_{f,r}^i \gets \text{FLIP}(x_r^i)$   \Comment{Based on Eq. (6)}
        \State $e_l^i \gets f(x_l^i;\theta)$
        \State $e_r^i \gets f(x_r^i;\theta)$
        \State $e_{f,l}^i \gets f(x_{f,l}^i;\theta)$
        \State $e_{f,r}^i \gets f(x_{f,r}^i;\theta)$
        \State Calculate $\mathcal{L}_{CE}$
        \State Calculate $\mathcal{L}_{\text{CC}}^{l\_{fr}},\mathcal{L}_{\text{CC}}^{{{fl}\_r}}, \mathcal{L}_{\text{CC}}^{r\_l}, \mathcal{L}_{\text{CC}}^{fr\_{fl}}$ \\
        \Comment{Based on Eqs. (9)-(12)}
        \State $\mathcal{L} \gets w_{ce} \times \mathcal{L}_{CE} + w_{cc} \times \mathcal{L}_{CC}$ \\
        \Comment{Based on Eq. (14)}

        \State Back-propagation and update $\theta$ based on $\mathcal{L}$
    \EndFor
\EndFor
\State \Return $\theta$
\EndFunction
\end{algorithmic}
\end{algorithm}

\begin{table*}[]
\centering
\caption{The cross-chirality experiments (L$\rightarrow$R) on public datasets.}
\resizebox{\textwidth}{!}{%
\begin{tabular}{llcccccccccc}
\hline
\multirow{2}{*}{Backbone}               & \multirow{2}{*}{Framework}   & \multicolumn{2}{c}{Red} & \multicolumn{2}{c}{Green} & \multicolumn{2}{c}{Blue} & 
                 \multicolumn{2}{c}{NIR} & \multicolumn{2}{c}{Tongji} \\ \cmidrule(lr){3-4} \cmidrule(lr){5-6} \cmidrule(lr){7-8} \cmidrule(lr){9-10} \cmidrule(lr){11-12}
             &     & ACC(\%)$\uparrow$        & EER(\%)$\downarrow$       & ACC(\%)$\uparrow$           & EER(\%)$\downarrow$         & ACC(\%)$\uparrow$           & EER(\%)$\downarrow$         & ACC(\%)$\uparrow$           & EER(\%)$\downarrow$        & ACC(\%)$\uparrow$            & EER(\%)$\downarrow$         \\ \hline
Shallow CNN~\cite{kumar2016identifying}& LRPR&             95.00&           12.7852&              95.45&            10.9778&             96.22&            9.0678&             94.22&           14.3730& 97.25        & 8.2089\\ \hline

\multirow{4}{*}{DenseNet101\cite{xu2022method}} & Naive &             77.56&           29.6444&              82.13&            25.7638&             77.88&            30.8152&             76.00&           31.2740& 80.74        & 26.9059\\
&LRPR&             86.70&           22.0889&              88.12&            20.6020&             87.85&            20.2983&             84.79&           23.1407& 88.39        & 19.4400\\

&\cellcolor{lightergray}CCPV &\cellcolor{lightergray}  97.92&\cellcolor{lightergray}  7.5259&\cellcolor{lightergray}  97.14&\cellcolor{lightergray}  8.7246 &\cellcolor{lightergray} 98.62       &\cellcolor{lightergray} 5.7457 &\cellcolor{lightergray}  97.09      &\cellcolor{lightergray}  8.8592 &\cellcolor{lightergray}  96.12 &\cellcolor{lightergray}  10.4133\\
&\cellcolor{lightergray}$\triangle$ &\cellcolor{lightergray}\textcolor{red}{+11.22}&\cellcolor{lightergray}\textcolor{red}{-14.5630}&\cellcolor{lightergray}\textcolor{red}{+9.02}&\cellcolor{lightergray}\textcolor{red}{-11.8774} &\cellcolor{lightergray}\textcolor{red}{+10.77}       &\cellcolor{lightergray}\textcolor{red}{-14.5526} &\cellcolor{lightergray}\textcolor{red}{+12.30}      &\cellcolor{lightergray}\textcolor{red}{-14.2815} &\cellcolor{lightergray}\textcolor{red}{+7.73} &\cellcolor{lightergray}\textcolor{red}{-9.0267}
\\ \hline

\multirow{4}{*}{ResNet18~\cite{he2016deep}} &Naive    & 74.62 & 32.9181 & 78.02 & 30.4592 & 74.23 & 33.0464 & 77.00 & 30.4323 & 82.69 & 26.3600 \\
&LRPR& 90.52 & 17.0962 & 89.52 & 18.1629 & 94.33 & 12.6962 & 88.92 & 18.5037 & 94.38 & 12.5151\\

&\cellcolor{lightergray}CCPV &\cellcolor{lightergray}  99.93&\cellcolor{lightergray}  1.1556&\cellcolor{lightergray}  99.91&\cellcolor{lightergray}  0.1740 &\cellcolor{lightergray} 99.99       &\cellcolor{lightergray} 0.3578 &\cellcolor{lightergray}  99.90      &\cellcolor{lightergray}  1.2296 &\cellcolor{lightergray}  99.93 &\cellcolor{lightergray}  1.2413\\
&\cellcolor{lightergray}$\triangle$ &\cellcolor{lightergray}\textcolor{red}{+9.41}&\cellcolor{lightergray}\textcolor{red}{-15.9406}&\cellcolor{lightergray}\textcolor{red}{+10.39}&\cellcolor{lightergray}\textcolor{red}{-17.9889} &\cellcolor{lightergray}\textcolor{red}{+5.66}       &\cellcolor{lightergray}\textcolor{red}{-12.3384} &\cellcolor{lightergray}\textcolor{red}{+10.98}      &\cellcolor{lightergray}\textcolor{red}{-17.2741} &\cellcolor{lightergray}\textcolor{red}{+5.55} &\cellcolor{lightergray}\textcolor{red}{-11.2738}
\\  \hline

\multirow{4}{*}{CompNet~\cite{liang2021compnet}}&Naive    & 76.49 & 30.2362 & 78.28 & 29.0963 & 75.90 & 30.3851 & 76.43 & 30.4148 & 85.86 & 20.8311 \\
&LRPR& 92.07 & 14.8876 & 91.49 & 15.2580 & 98.02 & 5.9556 & 91.07 & 14.6518 & 99.06 & 3.5556\\

&\cellcolor{lightergray}CCPV &\cellcolor{lightergray}  99.99&\cellcolor{lightergray}  0.1492&\cellcolor{lightergray}  99.99&\cellcolor{lightergray}  0.3259 &\cellcolor{lightergray} 99.99       &\cellcolor{lightergray} 0.1925 &\cellcolor{lightergray}  99.99      &\cellcolor{lightergray}  0.1881 &\cellcolor{lightergray}  99.99 &\cellcolor{lightergray}  0.3105\\
&\cellcolor{lightergray}$\triangle$ &\cellcolor{lightergray}\textcolor{red}{+7.92}&\cellcolor{lightergray}\textcolor{red}{-14.7384}&\cellcolor{lightergray}\textcolor{red}{+8.50}&\cellcolor{lightergray}\textcolor{red}{-14.9321} &\cellcolor{lightergray}\textcolor{red}{+1.97}       &\cellcolor{lightergray}\textcolor{red}{-5.7631} &\cellcolor{lightergray}\textcolor{red}{+8.63}      &\cellcolor{lightergray}\textcolor{red}{-14.4637} &\cellcolor{lightergray}\textcolor{red}{+0.93} &\cellcolor{lightergray}\textcolor{red}{-3.2451}
\\
\hline

\multirow{4}{*}{CO3Net~\cite{yang2023co}} &Naive    & 76.61 & 30.3234 & 75.85 & 30.9443 & 74.47 & 32.0148 & 77.59 & 29.4222 & 84.62 & 22.1556\\
&LRPR& 90.72 & 15.5407 & 90.93 & 15.2000 & 98.48 & 4.4889 & 91.36 & 15.3629 & 98.58 & 4.3511\\

&\cellcolor{lightergray}CCPV &\cellcolor{lightergray}  99.99&\cellcolor{lightergray}  0.1357&\cellcolor{lightergray}  99.97&\cellcolor{lightergray}  0.7407 &\cellcolor{lightergray} 99.99       &\cellcolor{lightergray} 0.1235 &\cellcolor{lightergray}  99.99      &\cellcolor{lightergray}  0.3259 &\cellcolor{lightergray}  99.97 &\cellcolor{lightergray}  0.4033\\
&\cellcolor{lightergray}$\triangle$ &\cellcolor{lightergray}\textcolor{red}{+9.27}&\cellcolor{lightergray}\textcolor{red}{-15.4050}&\cellcolor{lightergray}\textcolor{red}{+9.04}&\cellcolor{lightergray}\textcolor{red}{-14.4593} &\cellcolor{lightergray}\textcolor{red}{+1.51}       &\cellcolor{lightergray}\textcolor{red}{-4.3654} &\cellcolor{lightergray}\textcolor{red}{+8.63}      &\cellcolor{lightergray}\textcolor{red}{-15.0370} &\cellcolor{lightergray}\textcolor{red}{+1.39} &\cellcolor{lightergray}\textcolor{red}{-3.9478}
\\
\hline
\multirow{4}{*}{SACNet~\cite{gao2024scale}}&Naive   & 74.01 & 33.1492 & 75.01 & 32.4124 & 73.43 & 33.3410 & 74.87 & 32.5977 & 85.17 & 21.7600\\
&LRPR& 91.41 & 16.0296 & 89.78 & 17.3269 & 98.57 & 4.7556 & 92.02 & 16.9333 & 99.01 & 3.7733\\

&\cellcolor{lightergray}CCPV &\cellcolor{lightergray}  99.99&\cellcolor{lightergray}  0.1572&\cellcolor{lightergray}  99.98&\cellcolor{lightergray}  0.3703 &\cellcolor{lightergray} 99.99       &\cellcolor{lightergray} 0.3526 &\cellcolor{lightergray}  99.99      &\cellcolor{lightergray}  0.2719 &\cellcolor{lightergray}  99.99 &\cellcolor{lightergray}  0.8367\\
&\cellcolor{lightergray}$\triangle$ &\cellcolor{lightergray}\textcolor{red}{+8.58}&\cellcolor{lightergray}\textcolor{red}{-15.8724}&\cellcolor{lightergray}\textcolor{red}{+10.20}&\cellcolor{lightergray}\textcolor{red}{-16.9566} &\cellcolor{lightergray}\textcolor{red}{+1.42}       &\cellcolor{lightergray}\textcolor{red}{-4.4030} &\cellcolor{lightergray}\textcolor{red}{+7.97}      &\cellcolor{lightergray}\textcolor{red}{-16.6614} &\cellcolor{lightergray}\textcolor{red}{+0.98} &\cellcolor{lightergray}\textcolor{red}{-2.9366}
\\
\hline
\multirow{4}{*}{CCNet~\cite{yang2023comprehensive}}&Naive    & 77.09 & 29.7629 & 77.49 & 29.5259 & 73.87 & 33.6076 & 77.05 & 30.0148 & 85.17 & 21.9136 \\
&LRPR& 92.14 & 14.6074 & 89.74 & 15.6889 & 98.60 & 5.0667 & 90.96 & 15.5851 & 99.01 & 3.7733\\

&\cellcolor{lightergray}CCPV &\cellcolor{lightergray}  99.99&\cellcolor{lightergray}  0.1185&\cellcolor{lightergray}  99.99&\cellcolor{lightergray}  0.3703 &\cellcolor{lightergray} 99.99       &\cellcolor{lightergray} 0.3526 &\cellcolor{lightergray}  99.99      &\cellcolor{lightergray}  0.2370 &\cellcolor{lightergray}  99.99 &\cellcolor{lightergray}  0.1800\\
&\cellcolor{lightergray}$\triangle$ &\cellcolor{lightergray}\textcolor{red}{+7.85}&\cellcolor{lightergray}\textcolor{red}{-14.4889}&\cellcolor{lightergray}\textcolor{red}{+10.25}&\cellcolor{lightergray}\textcolor{red}{-15.3186} &\cellcolor{lightergray}\textcolor{red}{+1.39}       &\cellcolor{lightergray}\textcolor{red}{-4.7141} &\cellcolor{lightergray}\textcolor{red}{+9.03}      &\cellcolor{lightergray}\textcolor{red}{-15.3481} &\cellcolor{lightergray}\textcolor{red}{+0.98} &\cellcolor{lightergray}\textcolor{red}{-3.5933}
\\ \hline
\end{tabular}
}
\label{tab:1}
\end{table*}

\begin{table*}[]
\centering
\caption{The cross-chirality experiments (R$\rightarrow$L) on public datasets.}
\resizebox{\textwidth}{!}{%
\begin{tabular}{llcccccccccc}
\hline
\multirow{2}{*}{Backbone}               & \multirow{2}{*}{Framework}   & \multicolumn{2}{c}{Red} & \multicolumn{2}{c}{Green} & \multicolumn{2}{c}{Blue} & 
                 \multicolumn{2}{c}{NIR} & \multicolumn{2}{c}{Tongji} \\ \cmidrule(lr){3-4} \cmidrule(lr){5-6} \cmidrule(lr){7-8} \cmidrule(lr){9-10} \cmidrule(lr){11-12}
             &     & ACC(\%)$\uparrow$        & EER(\%)$\downarrow$       & ACC(\%)$\uparrow$           & EER(\%)$\downarrow$         & ACC(\%)$\uparrow$           & EER(\%)$\downarrow$         & ACC(\%)$\uparrow$           & EER(\%)$\downarrow$        & ACC(\%)$\uparrow$            & EER(\%)$\downarrow$         \\ \hline
Shallow CNN~\cite{kumar2016identifying}& LRPR&             95.02&           12.2361&              72.00&            10.8296&             97.14&            8.4148&             94.55&           13.3479& 97.86        & 7.2222\\ \hline

\multirow{4}{*}{DenseNet101\cite{xu2022method}} & Naive &             77.89&           29.4127&              81.96&            25.7541&             77.55&            30.8967&             76.07&           31.1120& 80.75        & 26.8844\\
&LRPR&             85.77&           22.3407&              87.62&            20.5551&             87.53&            20.9228&             83.13&           24.3851& 89.07        & 19.3648\\

&\cellcolor{lightergray}CCPV &\cellcolor{lightergray}  97.92&\cellcolor{lightergray}  7.5146&\cellcolor{lightergray}  97.19&\cellcolor{lightergray}  8.6167 &\cellcolor{lightergray} 98.52       &\cellcolor{lightergray} 5.9851 &\cellcolor{lightergray}  99.90      &\cellcolor{lightergray}  9.1407 &\cellcolor{lightergray}  96.09 &\cellcolor{lightergray}  10.4776\\
&\cellcolor{lightergray}$\triangle$ &\cellcolor{lightergray}\textcolor{red}{+12.15}&\cellcolor{lightergray}\textcolor{red}{-14.8261}&\cellcolor{lightergray}\textcolor{red}{+9.57}&\cellcolor{lightergray}\textcolor{red}{-11.9384} &\cellcolor{lightergray}\textcolor{red}{+10.99}       &\cellcolor{lightergray}\textcolor{red}{-14.9377} &\cellcolor{lightergray}\textcolor{red}{+16.77}      &\cellcolor{lightergray}\textcolor{red}{-15.2444} &\cellcolor{lightergray}\textcolor{red}{+7.02} &\cellcolor{lightergray}\textcolor{red}{-8.8872}
\\ \hline

\multirow{4}{*}{ResNet18~\cite{he2016deep}} &Naive    & 74.76 & 32.5333 & 77.93 & 30.3981 & 74.12 & 33.2839 & 76.57 & 30.6954 & 82.93 & 26.2622 \\
&LRPR& 88.95 & 18.0148 & 87.78 & 19.3629 & 95.67 & 11.1259 & 87.12 & 20.3733 & 94.03 & 13.2444\\

&\cellcolor{lightergray}CCPV &\cellcolor{lightergray}  99.91&\cellcolor{lightergray}  1.2592&\cellcolor{lightergray}  99.89&\cellcolor{lightergray}  1.4074 &\cellcolor{lightergray} 99.99       &\cellcolor{lightergray} 0.5339 &\cellcolor{lightergray}  99.98      &\cellcolor{lightergray}  1.5111 &\cellcolor{lightergray}  99.82 &\cellcolor{lightergray}  1.3100\\
&\cellcolor{lightergray}$\triangle$ &\cellcolor{lightergray}\textcolor{red}{+10.96}&\cellcolor{lightergray}\textcolor{red}{-16.7556}&\cellcolor{lightergray}\textcolor{red}{+12.11}&\cellcolor{lightergray}\textcolor{red}{-17.9555} &\cellcolor{lightergray}\textcolor{red}{+4.32}       &\cellcolor{lightergray}\textcolor{red}{-10.5920} &\cellcolor{lightergray}\textcolor{red}{+12.86}      &\cellcolor{lightergray}\textcolor{red}{-18.8622} &\cellcolor{lightergray}\textcolor{red}{+5.79} &\cellcolor{lightergray}\textcolor{red}{-11.9344}
\\  \hline

\multirow{4}{*}{CompNet~\cite{liang2021compnet}}&Naive    & 76.30 & 30.4296 & 78.17 & 29.3704 & 75.61 & 30.5703 & 76.85 & 31.5703 & 84.70 & 20.9874 \\
&LRPR& 90.73 & 15.3037 & 91.49 & 15.1296 & 97.99 & 5.7407 & 91.72 & 14.7259 & 99.01 & 3.4708\\

&\cellcolor{lightergray}CCPV &\cellcolor{lightergray}  99.99&\cellcolor{lightergray}  0.1492&\cellcolor{lightergray}  99.96&\cellcolor{lightergray}  0.3703 &\cellcolor{lightergray} 99.99       &\cellcolor{lightergray} 0.1844 &\cellcolor{lightergray}  99.99      &\cellcolor{lightergray}  0.2815 &\cellcolor{lightergray}  99.96 &\cellcolor{lightergray}  0.4804\\
&\cellcolor{lightergray}$\triangle$ &\cellcolor{lightergray}\textcolor{red}{+9.26}&\cellcolor{lightergray}\textcolor{red}{-15.1545}&\cellcolor{lightergray}\textcolor{red}{+8.47}&\cellcolor{lightergray}\textcolor{red}{-14.7593} &\cellcolor{lightergray}\textcolor{red}{+2.00}       &\cellcolor{lightergray}\textcolor{red}{-5.5563} &\cellcolor{lightergray}\textcolor{red}{+8.27}      &\cellcolor{lightergray}\textcolor{red}{-14.4444} &\cellcolor{lightergray}\textcolor{red}{+0.95} &\cellcolor{lightergray}\textcolor{red}{-2.9904}
\\
\hline

\multirow{4}{*}{CO3Net~\cite{yang2023co}} &Naive    & 74.69 & 32.5037 & 74.93 & 32.2277 & 73.87 & 33.8704 & 74.63 & 33.6537 & 84.74 & 20.4741\\
&LRPR& 92.17 & 16.4074 & 91.49 & 16.6296 & 98.59 & 5.3481 & 92.07 & 16.3556 & 99.09 & 3.0467\\

&\cellcolor{lightergray}CCPV &\cellcolor{lightergray}  99.99&\cellcolor{lightergray}  0.1629&\cellcolor{lightergray}  99.98&\cellcolor{lightergray}  0.4148 &\cellcolor{lightergray} 99.99       &\cellcolor{lightergray} 0.4511 &\cellcolor{lightergray}  99.99      &\cellcolor{lightergray}  0.3130 &\cellcolor{lightergray}  99.99 &\cellcolor{lightergray}  0.1315\\
&\cellcolor{lightergray}$\triangle$ &\cellcolor{lightergray}\textcolor{red}{+7.82}&\cellcolor{lightergray}\textcolor{red}{-16.2445}&\cellcolor{lightergray}\textcolor{red}{+8.49}&\cellcolor{lightergray}\textcolor{red}{-16.2148} &\cellcolor{lightergray}\textcolor{red}{+1.40}       &\cellcolor{lightergray}\textcolor{red}{-4.8970} &\cellcolor{lightergray}\textcolor{red}{+7.92}      &\cellcolor{lightergray}\textcolor{red}{-16.0426} &\cellcolor{lightergray}\textcolor{red}{+0.90} &\cellcolor{lightergray}\textcolor{red}{-2.9152}
\\
\hline
\multirow{4}{*}{SACNet~\cite{gao2024scale}}&Naive   & 73.26 & 33.6815 & 77.99 & 32.3415 & 74.12 & 32.5185 & 74.95 & 32.5185 & 85.65 & 21.4822\\
&LRPR& 92.28 & 13.2389 & 91.49 & 14.8444 & 97.52 & 5.5778 & 93.02 & 14.9333 & 99.07 & 2.8178\\

&\cellcolor{lightergray}CCPV &\cellcolor{lightergray}  99.99&\cellcolor{lightergray}  0.1132&\cellcolor{lightergray}  99.98&\cellcolor{lightergray}  0.2296 &\cellcolor{lightergray} 99.99       &\cellcolor{lightergray} 0.1496 &\cellcolor{lightergray}  99.99      &\cellcolor{lightergray}  0.1630 &\cellcolor{lightergray}  99.99 &\cellcolor{lightergray}  0.1804\\
&\cellcolor{lightergray}$\triangle$ &\cellcolor{lightergray}\textcolor{red}{+7.71}&\cellcolor{lightergray}\textcolor{red}{-13.1257}&\cellcolor{lightergray}\textcolor{red}{+8.49}&\cellcolor{lightergray}\textcolor{red}{-14.6148} &\cellcolor{lightergray}\textcolor{red}{+2.47}       &\cellcolor{lightergray}\textcolor{red}{-5.4282} &\cellcolor{lightergray}\textcolor{red}{+6.97}      &\cellcolor{lightergray}\textcolor{red}{-14.7703} &\cellcolor{lightergray}\textcolor{red}{+0.92} &\cellcolor{lightergray}\textcolor{red}{-2.6374}
\\
\hline
\multirow{4}{*}{CCNet~\cite{yang2023comprehensive}}&Naive    & 74.86 & 33.4426 & 77.22 & 31.6741 & 74.12 & 32.5185 & 74.95 & 30.3703& 85.65 & 21.4822 \\
&LRPR& 92.43 & 13.2889 & 91.78 & 14.8444 & 97.52 & 5.5778 & 93.02 & 14.9333 & 99.07 & 2.8178\\

&\cellcolor{lightergray}CCPV &\cellcolor{lightergray}  99.99&\cellcolor{lightergray}  0.1132&\cellcolor{lightergray}  99.98&\cellcolor{lightergray}  0.2296 &\cellcolor{lightergray} 99.99       &\cellcolor{lightergray} 0.1496 &\cellcolor{lightergray}  99.99      &\cellcolor{lightergray}  0.1630 &\cellcolor{lightergray}  99.99 &\cellcolor{lightergray}  0.1804\\
&\cellcolor{lightergray}$\triangle$ &\cellcolor{lightergray}\textcolor{red}{+6.71}&\cellcolor{lightergray}\textcolor{red}{-13.1757}&\cellcolor{lightergray}\textcolor{red}{+8.43}&\cellcolor{lightergray}\textcolor{red}{-14.6918} &\cellcolor{lightergray}\textcolor{red}{+0.67}       &\cellcolor{lightergray}\textcolor{red}{-3.4371} &\cellcolor{lightergray}\textcolor{red}{+8.37}      &\cellcolor{lightergray}\textcolor{red}{-14.7407} &\cellcolor{lightergray}\textcolor{red}{+0.63} &\cellcolor{lightergray}\textcolor{red}{-2.8811}
\\ \hline
\end{tabular}
}
\label{tab:2}
\end{table*}

\section{Experiments}
\subsection{Experimental Settings}
\subsubsection{Datasets}
This paper applies two public datasets containing left and right palms for experiments.

\textbf{Tongji}~\cite{zhang2017towards}: Tongji consists of 12000 images derived from 600 distinct palms using the proprietary touchless acquisition device. These images were collected from 300 volunteers at Tongji University, including both left and right palmprints. The data collection occurred over two sessions, each capturing 10 images per palm, resulting in 20 images per individual. 

\textbf{Multi-Spectral}~\cite{zhang2009online}: Multi-Spectral contains 6000 images obtained from 250 volunteers. This Multi-Spectral dataset includes cropped palmprint images across four spectral bands: Red, Green, Blue, and NIR. The images were collected in two sessions; six images per palm were captured during each session, resulting in 12 images per person. Thus, 48 spectral images were collected from each volunteer.

\begin{figure*}
  \begin{minipage}[t]{0.195\linewidth}
  \centering
  \includegraphics[width=\textwidth]{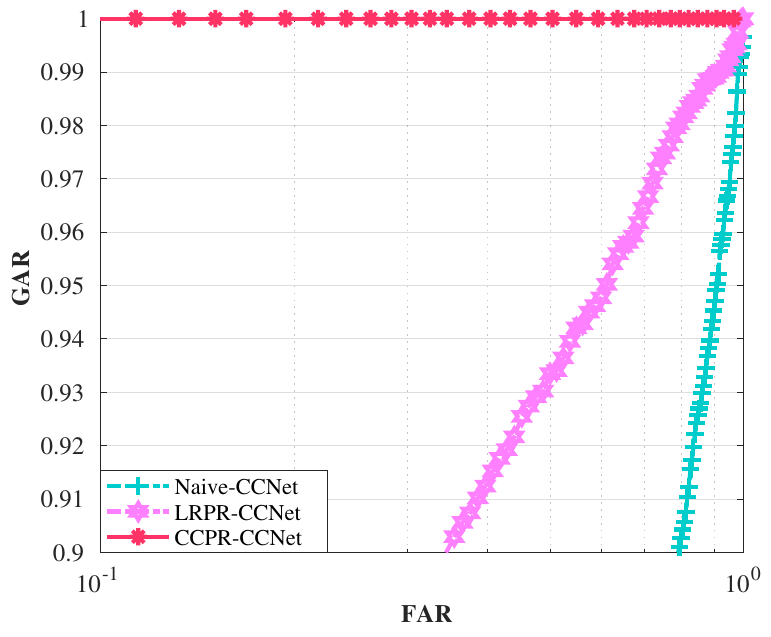}
   \centerline{(a)}
   \end{minipage}  
  \begin{minipage}[t]{0.195\linewidth}
  \centering
  \includegraphics[width=\textwidth]{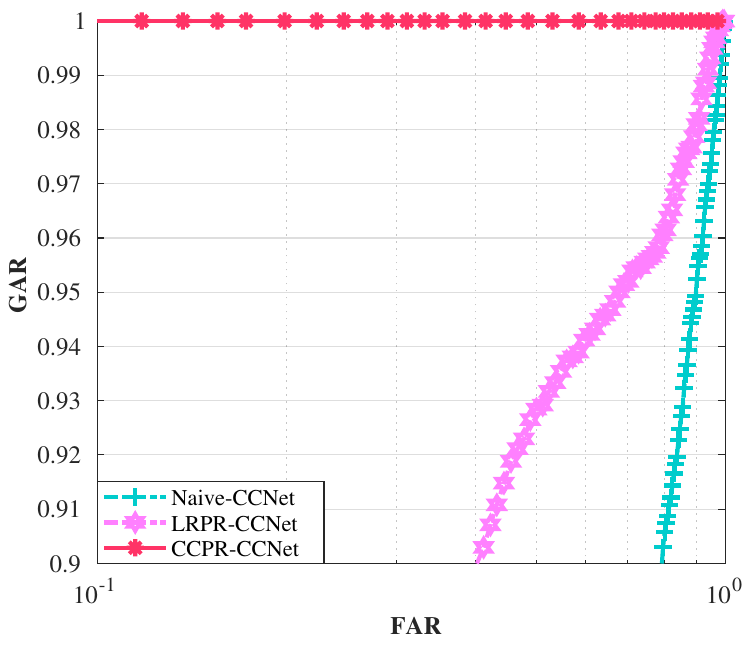}
   \centerline{(b)}
   \end{minipage}
   \begin{minipage}[t]{0.195\linewidth}
  \centering
  \includegraphics[width=\textwidth]{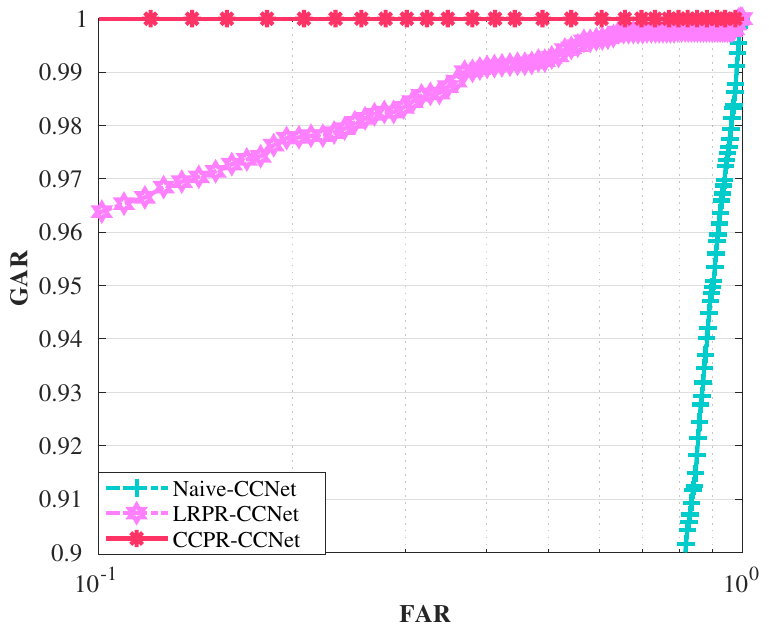}
   \centerline{(c)}
   \end{minipage}
   \begin{minipage}[t]{0.195\linewidth}
  \centering
  \includegraphics[width=\textwidth]{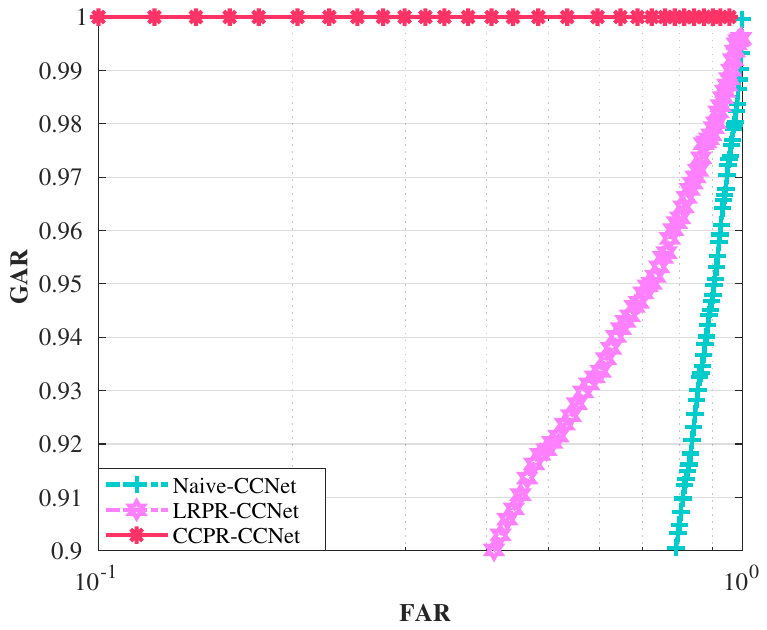}
   \centerline{(d)}
   \end{minipage}
   \begin{minipage}[t]{0.195\linewidth}
  \centering
  \includegraphics[width=\textwidth]{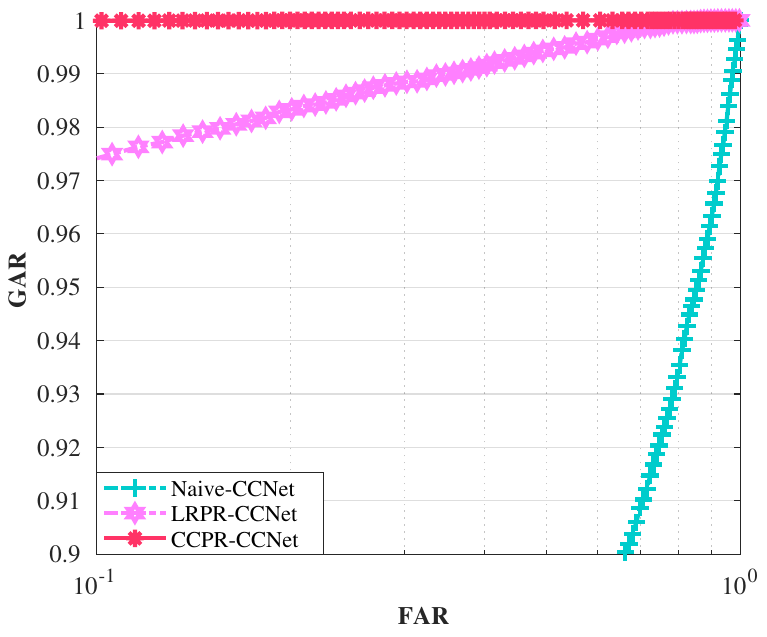}
   \centerline{(e)}
   \end{minipage}
   \\
     \begin{minipage}[t]{0.195\linewidth}
  \centering
  \includegraphics[width=\textwidth]{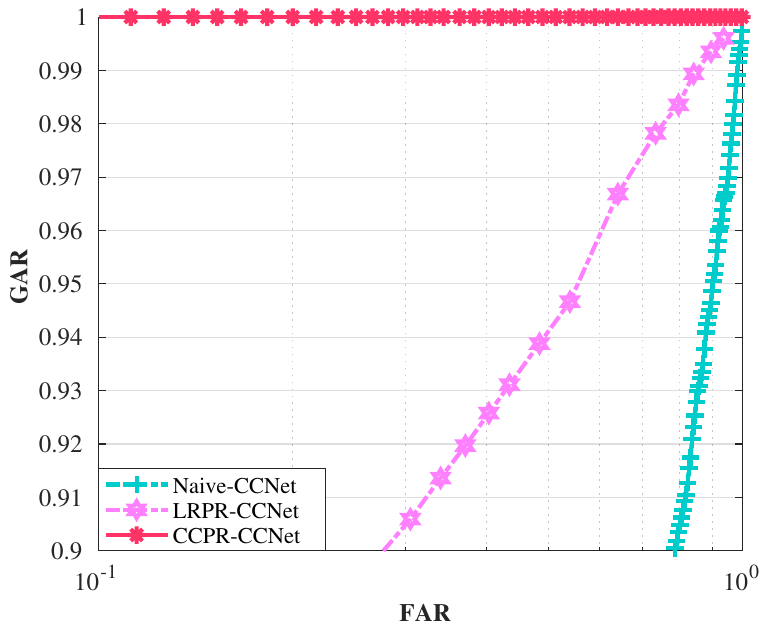}
   \centerline{(f)}
   \end{minipage}  
  \begin{minipage}[t]{0.195\linewidth}
  \centering
  \includegraphics[width=\textwidth]{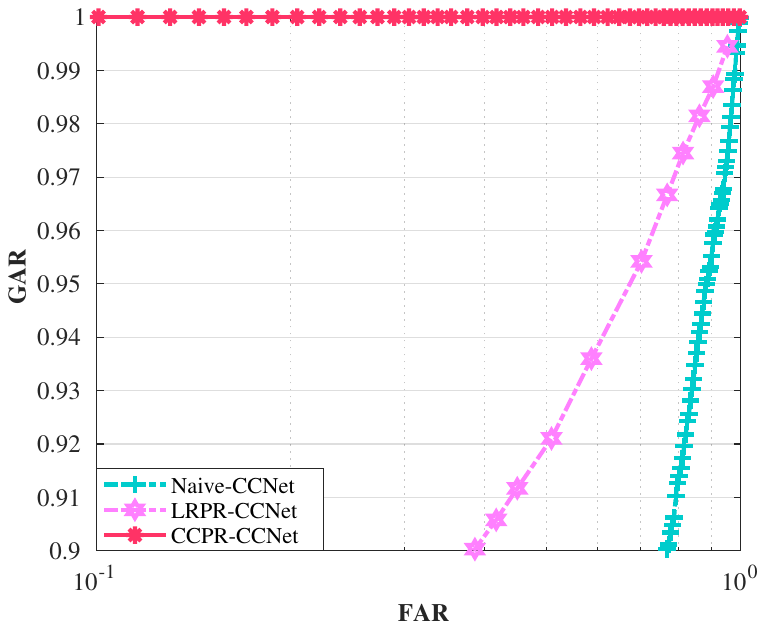}
   \centerline{(g)}
   \end{minipage}
   \begin{minipage}[t]{0.195\linewidth}
  \centering
  \includegraphics[width=\textwidth]{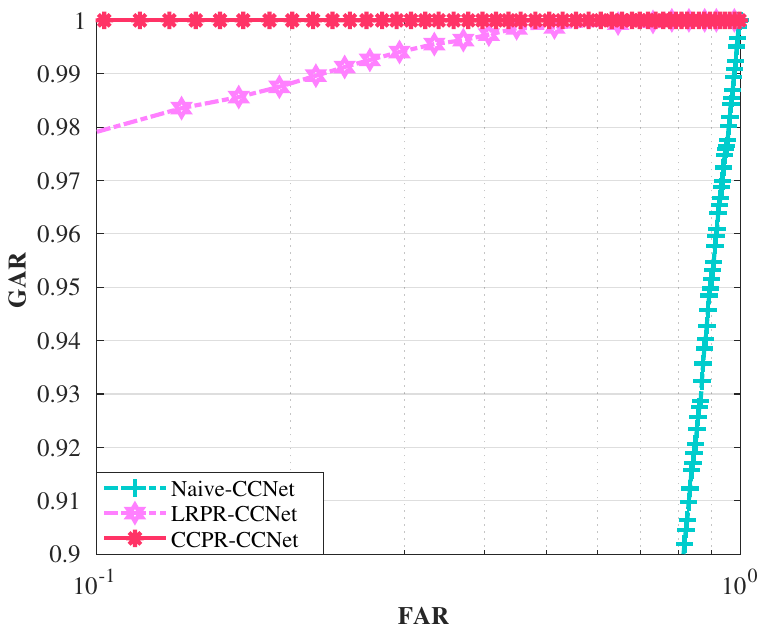}
   \centerline{(h)}
   \end{minipage}
   \begin{minipage}[t]{0.195\linewidth}
  \centering
  \includegraphics[width=\textwidth]{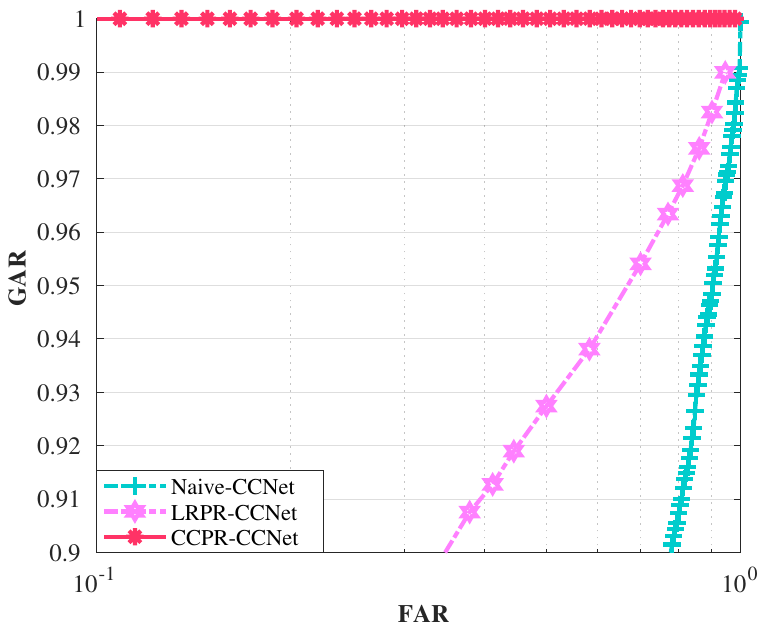}
   \centerline{(i)}
   \end{minipage}
   \begin{minipage}[t]{0.195\linewidth}
  \centering
  \includegraphics[width=\textwidth]{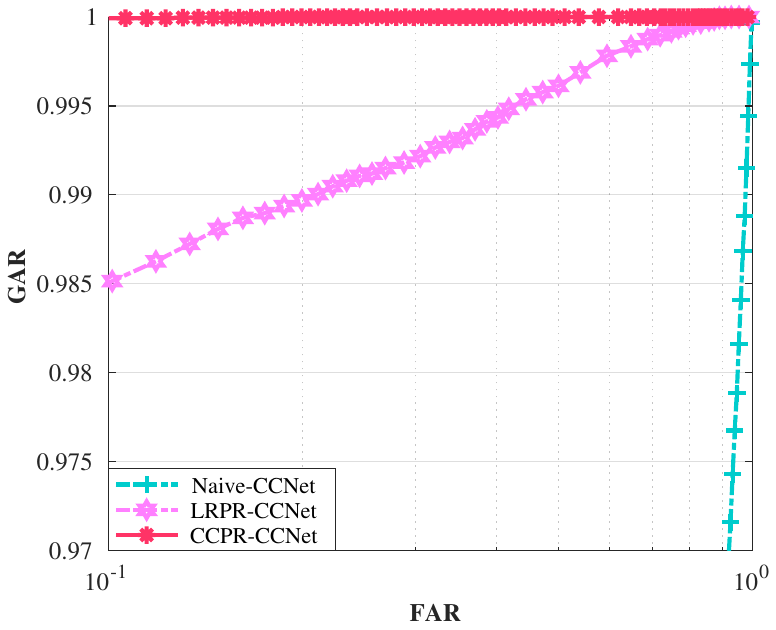}
   \centerline{(j)}
   \end{minipage} \\
    
     \caption{The ROC curves of the proposed CCPV and its competing methods across Red, Green, Blue, NIR, and Tongji datasets. (a)-(e) denote the L$\rightarrow$R ROC curves. (f)-(j) denote the R$\rightarrow$L ROC curves.}
  \label{fig:roc}
\end{figure*}

\subsubsection{Implementation Details}
The CCPV is implemented with the PyTorch framework, with optimization performed by the Adam optimizer~\cite{kingma2014adam} (set learning rate as 0.001). The batch size and training epoch are 512 and 500, respectively. The experiments are conducted on a single NVIDIA RTX 3090 GPU.

In the Tongji dataset, the training set consists of 2,400 palmprint images from 300 identities, with five left and five right palmprint images per identity. The remaining 9,600 palmprint images from the same 300 identities, with 15 left and 15 right palmprint images per identity, are reserved for testing.

In the Multi-Spectral dataset, for each spectrum, the training set comprises 1,200 palmprint images from 250 identities, with three left palmprint images and three right palmprint images per identity. The test set includes nine left and nine right palmprint images per identity.

\subsubsection{Backbones}
We select two representative networks, ResNet18~\cite{he2016deep} and DenseNet101~\cite{xu2022method}. Besides, several typical deep-learning-based palmprint recognition models, including CompNet~\cite{liang2021compnet}, CO3Net~\cite{yang2023co}, SACNet~\cite{gao2024scale}, and CCNet~\cite{yang2023comprehensive}, are selected to validate the performance.

\subsubsection{Training Frameworks}

Traditional palmprint recognition systems, referred to as the \textit{traditional framework}, typically train models using either left or right palmprints exclusively. To address cross-chirality palmprint verification while maintaining a fair comparative basis, we propose a \textit{naive framework}. This approach involves assigning identical labels to each individual's left and right palmprints and utilizing both for model training. This method represents a straightforward extension of conventional techniques to accommodate cross-chirality verification.

In addition to the vanilla and naive frameworks, we include a third training framework in our comparison: LRPR, proposed by Kumar~\textit{et al.}~\cite{kumar2016identifying}. The LRPR framework uses structural similarity between left and right palmprints by flipping. This method provides an alternative for cross-chirality verification, distinct from vanilla and naive frameworks.

\subsubsection{Metrics}
We assess verification performance using Receiver Operating Characteristic (ROC) curves and the Equal Error Rate (EER), which are both derived from the Genuine Acceptance Rate (GAR) and False Acceptance Rate (FAR). Additionally, GAR@0.1\%FAR and GAR@0.0001\%FAR represent the Genuine Acceptance Rate at False Acceptance Rates of 0.1\% and 0.0001\%, respectively. The accuracy (ACC) metric is also calculated to evaluate the system's identification performance.

\begin{table*}[]
\centering
\caption{Comparison of the GAR@0.001\%FAR indicator of cross left-right palmprint matching between CCNet and CCPV-CCNet on public datasets.}
\begin{tabular}{ccccccccc}
\hline
\multirow{2}{*}{Frameworks} & \multicolumn{3}{c}{Experiment settings} & \multicolumn{5}{c}{Datasets} \\ 
\cmidrule(lr){2-4} \cmidrule(lr){5-9}
& Train & Gallery & Query & Red & Green & Blue & NIR & Tongji \\ \hline

\multirow{4}{*}{Traditional-CCNet} 
& L & L & L & \textbf{99.99} & \textbf{99.99} & \textbf{99.99} & \textbf{99.99} & \textbf{99.99} \\
& R& R& R& \textbf{99.99} & \textbf{99.99} & \textbf{99.99} & \textbf{99.99 }& \textbf{99.99} \\
& L & L & R& 0.33& 0.24& 0.03& 0.35& 0.25\\
& R& R& L & 0.03& 0.51& 0.03& 0.45& 0.12\\ \hline

\multirow{4}{*}{Naive-CCNet} 
& L+R& L& L& \textbf{99.99} & \textbf{99.99} & \textbf{99.99 }& \textbf{99.99} & 99.93 \\
& L+R& R& R& 94.31 & 94.31 & 94.27 & 94.34 & 99.72\\
& L+R& L & R& 29.76 & 37.19 & 33.84 & 35.26 & 67.20 \\
& L+R& R& L & 35.35 & 36.95 & 33.72 & 35.29 & 67.08 \\ \hline

\multirow{4}{*}{CCPV-CCNet} 
& L+R& L & L & 99.87 & 99.90 & \textbf{99.99} & 99.59 & 99.76 \\
& L+R& R& R& 99.81& 99.79& \textbf{99.99}& 99.53& 99.69 \\
& L+R& L & R& \textbf{99.85} & \textbf{99.80} & \textbf{99.98} & \textbf{99.58} & \textbf{99.76} \\
& L+R& R& L & \textbf{99.87} & \textbf{99.80} & \textbf{99.97} & \textbf{99.55} & \textbf{99.68} \\ \hline
\end{tabular}
\label{tab:3}
\vspace{2pt}
\parbox{\textwidth}{\centering ``Naive'' uses both left and right palmprints during the training phase.}
\end{table*}

\begin{table*}[!t]
\centering
\caption{EERs (\%) are under different spectrum pairings for three frameworks.}
\resizebox{.99\textwidth}{!}{%
\begin{tabular}{ccccccccccccc}
\hline
\multirow{2}{*}{Datasets} & \multicolumn{4}{c}{Naive-CCNet} & \multicolumn{4}{c}{LRPR-CCNet} & \multicolumn{4}{c}{CCPV-CCNet} \\ 
\cmidrule(lr){2-5} \cmidrule(lr){6-9} \cmidrule(lr){10-13}
& Red & Green & Blue & NIR & Red & Green & Blue & NIR & Red & Green & Blue & NIR \\ \hline
Red  & 29.7629& 33.6458& 35.6107& 36.9265& 14.6074& 16.0000& 16.5333& 16.7508& 0.1185& 1.0667& 2.8445& 1.0660\\
Green & 31.3037& 29.5259& 30.2427& 39.8370& 15.5851& 15.6889& 14.3179& 18.4296& 0.9776& 0.1481& 2.8444& 3.1407\\
Blue  & 39.3481& 36.9217& 33.6076& 45.9480& 4.0991& 4.7259& 5.0667& 6.4000& 2.0881& 1.3185& 0.0444& 2.9185\\
NIR     & 31.6889& 36.1877& 39.4667& 30.0148& 15.3724& 18.3139& 19.1408& 15.5851& 1.1672& 3.0962& 3.1551& 0.2370\\
\hline
\end{tabular}
}
\label{tab:4}
\end{table*}

\begin{table*}[!t]
\centering
\caption{Performance Evaluation Across Different Datasets (Tongji$\rightarrow$Multi-Spectral) for three frameworks.}
\resizebox{.99\textwidth}{!}{%
\begin{tabular}{cccccccccc}
\hline
\multirow{2}{*}{Datasets} & \multicolumn{3}{c}{Naive-CCNet} & \multicolumn{3}{c}{LRPR-CCNet} & \multicolumn{3}{c}{CCPV-CCNet} \\ 
\cmidrule(lr){2-4} \cmidrule(lr){5-7} \cmidrule(lr){8-10}
& GAR@0.1\%FAR & ACC   & EER   & GAR@0.1\%FAR & ACC   & EER   & GAR@0.1\%FAR & ACC   & EER   \\ \hline
Tongji$\rightarrow$Red    & 31.88& 60.43& 44.02& 73.08& 83.01& 22.0592& 95.20        & 95.93 & 5.3000\\
Tongji$\rightarrow$Green  & 31.64& 61.17& 43.9851& 73.40& 82.56& 22.2799& 95.08        & 95.89 & 5.5111\\
Tongji$\rightarrow$Blue   & 31.93& 59.05& 49.6272& 93.40& 97.19& 8.3631& 95.28        & 97.71 & 5.3333\\
Tongji$\rightarrow$NIR    & 33.79& 63.30& 42.5976& 75.88& 85.66& 20.7703& 95.41        & 96.49 & 5.2778\\
\hline
\end{tabular}
}
\label{tab:5}
\end{table*}

\begin{table*}[!t]
\centering
\caption{Performance Evaluation Across Different Datasets (Multi-Spectral$\rightarrow$Tongji) for three frameworks.}
\resizebox{.99\textwidth}{!}{%
\begin{tabular}{cccccccccc}
\hline
\multirow{2}{*}{Datasets} & \multicolumn{3}{c}{Naive-CCNet} & \multicolumn{3}{c}{LRPR-CCNet} & \multicolumn{3}{c}{CCPV-CCNet} \\ 
\cmidrule(lr){2-4} \cmidrule(lr){5-7} \cmidrule(lr){8-10}
& GAR@0.1\%FAR & ACC   & EER   & GAR@0.1\%FAR & ACC   & EER   & GAR@0.1\%FAR & ACC   & EER   \\ \hline
Red$\rightarrow$Tongji& 50.39& 75.45& 31.7692& 83.53& 92.43& 14.6833& 92.13& 98.58& 4.8467\\
Green$\rightarrow$Tongji& 53.18& 76.84& 30.8394& 83.56& 92.14& 15.0367& 96.54& 98.49& 4.6433\\
Blue$\rightarrow$Tongji& 51.52& 77.12& 30.6800& 83.97& 92.73& 14.2800& 97.04& 98.75& 3.8933\\
NIR$\rightarrow$Tongji& 52.51& 77.82& 30.2310& 83.82& 92.77& 14.6333& 96.17& 98.33& 5.0167\\
\hline
\end{tabular}
}
\label{tab:6}
\end{table*}

\subsection{Cross-Chirality Palmprint Verification Experiments}
\subsubsection{Close-Set Experiments}

We began by conducting closed-set experiments on public datasets to assess cross-chirality performance. The results of different training frameworks are presented in Tabs.~\ref{tab:1} and~\ref{tab:2}. Table~\ref{tab:1} details the outcomes for left-to-right (L$\rightarrow$R) palmprint matching, while Table~\ref{tab:2} covers right-to-left (R$\rightarrow$L) palmprint matching.

The results reveal that networks trained with naive framework perform poorly. This is primarily due to these algorithms' tendency to mistakenly classify palmprints from different hands as coming from the same individual, due to a lack of heterogeneous feature constraints between left and right palms. Consequently, these models fail to effectively learn cross-chirality features. In contrast, the LRPR framework shows improvement by incorporating a flipping operation that uses prior knowledge to partially align left and right palmprint features.

Our proposed CCPV framework consistently delivers the best performance across various backbones. The significant performance differences indicated by the ``$\triangle$'' symbols in the tables highlight that CCPV markedly outperforms LRPR. This superiority is attributed to our matching rule and loss function, which effectively utilize the symmetrical properties of palmprints to reduce structural gaps and matching variance. Moreover, CCPV enhances the network's ability to extract cross-chirality features and create an optimal feature space.

Fig.~\ref{fig:roc} displays the ROC curves for the cross-chirality performance of different frameworks using CCNet. Our framework evidently achieves the highest GAR for a given FAR, thanks to the cross-chirality feature space created by our proposed CC Loss.

We also evaluated the performance of the proposed method with various matching pairs: L$\rightarrow$L, R$\rightarrow$R, L$\rightarrow$R, and R$\rightarrow$L. The GAR@0.001\%FAR metric was used for evaluation, with results shown in Table~\ref{tab:3}. For the traditional framework, which only uses unilateral palm prints during training, the results for L$\rightarrow$R and R$\rightarrow$L show a failure in cross-matching. 

While the performance gap between CCPV and the naive framework is minimal for L$\rightarrow$L and R$\rightarrow$R matching, the naive framework struggles with L$\rightarrow$R and R$\rightarrow$L matching. Despite being trained with both left and right palmprints, the naive approach lacks feature constraints between the two, leading to suboptimal verification. In contrast, the CCPV framework achieves superior performance, effectively enabling cross-chirality verification.

\begin{table}[!t]
\caption{EER (\%) under different matching rules for CCPV-CCNet.}
\centering
\resizebox{\columnwidth}{!}{
\begin{tabular}{lcccccc}
\hline
               & Red   & Green & Blue  & NIR   & Tongji & Average \\  \hline
Competition            & 0.2667& 0.3111& 0.0971& 0.4766& 0.2200& 0.2743\\
Ours           & \textbf{0.1185}& \textbf{0.1481}& \textbf{0.0444}& \textbf{0.2370}& \textbf{0.1800}& \textbf{0.1456}\\ 
\hline
\end{tabular}}
\label{tab:7}
\end{table}

\begin{table*}[!t]
\centering
\caption{Ablation Experiments about Loss Functions.}
\resizebox{.8\textwidth}{!}{%
\begin{tabular}{ccccccccccc}
\hline
\multicolumn{5}{c}{Loss} & \multicolumn{6}{c}{Datasets} \\ \cmidrule(lr){1-5} \cmidrule(lr){6-11} 
$\mathcal{L}_{\text{CC}}^{l\_fr}$ & $\mathcal{L}_{\text{CC}}^{{fl\_r}}$ & $\mathcal{L}_{\text{CC}}^{r\_l}$ & $\mathcal{L}_{\text{CC}}^{fr\_fl}$ & $\mathcal{L}_{\text{CE}}$ & Red                  & Green                & Blue                 & NIR                  & Tongji               & \textbf{Average}       \\ \hline
$\times$                          & $\times$                            & $\times$                         & $\times$                           & $\checkmark$              & \multicolumn{1}{c}{17.1852} & \multicolumn{1}{c}{17.7037} & \multicolumn{1}{c}{7.7481} & \multicolumn{1}{c}{17.5703} & \multicolumn{1}{c}{7.3467} & 13.1108 \\
$\checkmark$                      & $\checkmark$                        & $\checkmark$                     & $\checkmark$                       & $\times$                  & \multicolumn{1}{c}{0.3948} & \multicolumn{1}{c}{0.4444} & \multicolumn{1}{c}{0.3556} & \multicolumn{1}{c}{0.7493} & \multicolumn{1}{c}{0.6767} & 0.5242 \\
$\checkmark$                      & $\checkmark$                        & $\times$                         & $\times$                           & $\checkmark$              & 0.8592                & 0.4221                & 0.3407                & 0.8296                & 0.5500                & 0.6003 \\
$\times$                          & $\times$                            & $\checkmark$                     & $\checkmark$                       & $\checkmark$              & 0.4444                & 0.5333                & 0.2004                & 1.7925                & 0.3969                & 0.6735 \\
$\checkmark$                      & $\checkmark$                        & $\checkmark$                     & $\checkmark$                       & $\checkmark$              & \textbf{0.1185}       & \textbf{0.1481}       & \textbf{0.0444}       & \textbf{0.2370}       & \textbf{0.1800}       & \textbf{0.1456} \\ \hline
\end{tabular}
}
\label{tab:8}
\end{table*}

\subsubsection{Cross-spectral Experiments}

We further assessed the performance of cross-spectral palmprint recognition by using palmprint images from one spectrum as gallery samples and those from another spectrum as probe samples. We set the naive framework as baseline and compared our method against Naive-CCNet and LRPR-CCNet.

The results of these cross-spectral experiments are presented in Tabs.~\ref{tab:4}. The comparison of EERs across different frameworks clearly shows that our method significantly outperforms the baseline. To control for the influence of the recognition network, we utilized CCNet with the framework proposed in~\cite{kumar2016identifying}, with results also detailed in Table~\ref{tab:4}.

When comparing LRPR-CCNet to CCPV-CCNet, our method consistently achieves the lowest EER across all cross-spectral tests. This indicates that CCPV is more effective at distinguishing genuine matches from impostor matches in cross-spectral palmprint scenarios. Unlike LRPR-CCNet, our framework ensures intra-class feature consistency between left and right palmprints. It maintains similarity in intra-class features across different spectra, thus minimizing the feature gap between them.

\subsubsection{Open-set Experiments}
To further assess the effectiveness of our method, we conducted open-set palmprint verification experiments. We used the Tongji dataset for training and the multispectral dataset for testing. The results in Tabs.~\ref{tab:5} include EER and GAR@0.1\%FAR metrics. The data shows that CCPV-CCNet consistently outperforms Naive-CCNet across all metrics, demonstrating superior cross-dataset performance.

To validate CCPV's advantage over LRPR, we ensured consistency by using CCNet as the recognition network for both methods. The results indicate that CCPV-CCNet consistently surpasses LRPR-CCNet in all metrics, highlighting the efficacy of our approach in open-set scenarios. Our CCPV method performs admirably when transitioning from the Tongji dataset to the multispectral dataset.

Additionally, we reversed the roles by using the multispectral dataset for training and the Tongji dataset for testing. The results, shown in Tabs.~\ref{tab:6}, confirm that CCPV-CCNet continues to deliver the best performance, reinforcing the robustness and adaptability of our method.

\subsection{Ablation Study}
\subsubsection{The Matching Rule}
In Table~\ref{tab:7}, we present an ablation study evaluating the effectiveness of the matching rules within the CCPV framework using CCNet. This experiment compares our proposed four-matching rule against the traditional competition-matching rule across various datasets.

The competition matching rule, which selects the minimal matching result as the outcome, is commonly used in competitive methods~\cite{kong2004competitive}. Our ablation study assesses the performance of both matching rules, as shown in Table~\ref{tab:7}, by analyzing the EER across different datasets.

The results reveal that our four-matching rule significantly outperforms the competition-matching rule in all spectral conditions. This improvement underscores the rule's ability to reduce structural differences and variance at the matching level. The four-matching rule enhances recognition performance by aligning the matching strategy with the optimization constraints, leading to superior results.

\subsubsection{Loss Functions}
In Table~\ref{tab:8}, we present an ablation study to evaluate the impact of different loss functions on cross-chirality matching performance. This analysis reveals how each loss function contributes to the overall effectiveness of the CCPV-CCNet framework.

The results show that using only CE Loss, without CC Loss, results in the poorest cross-chirality verification performance, with an average EER of 13.1108\% across the five datasets. When only CC Loss is used, excluding CE Loss, CCPV-CCNet achieves an average EER of 0.5242\%. CE Loss is crucial for identifying clustering centers for left and right palmprints, while adding $\mathcal{L}_{\text{CC}}^{l\_fr}$ and $\mathcal{L}_{\text{CC}}^{{fl\_r}}$ to CE Loss reduces the EER by 12.5105\%. Similarly, incorporating $\mathcal{L}_{\text{CC}}^{r\_l}$ and $\mathcal{L}_{\text{CC}}^{fr\_fl}$ decreases the EER by 12.4373\%. 

The losses $\mathcal{L}_{\text{CC}}^{l\_fr}$ and $\mathcal{L}_{\text{CC}}^{{fl\_r}}$ constrain feature differences between left and flipped right palmprints, while $\mathcal{L}_{\text{CC}}^{r\_l}$ and $\mathcal{L}_{\text{CC}}^{fr\_fl}$ address feature differences between left and right palms. All four loss functions contribute positively to the validation experiments. Combining these losses in our proposed method demonstrates a clear superiority in improving cross-chirality verification performance.

\subsection{Visualization Analysis}
\begin{figure}
  \centering
  \includegraphics[width=0.99\columnwidth]{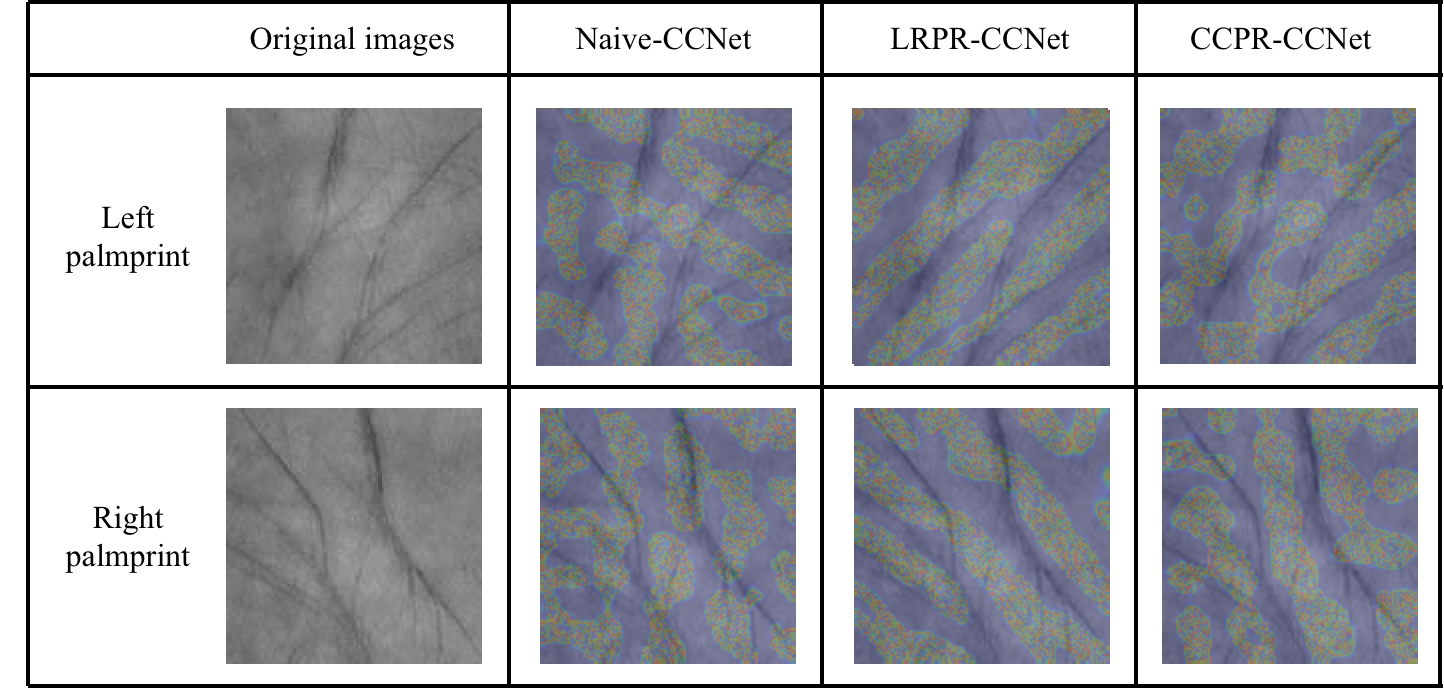}
     \caption{Sample palmprint images of the same identity from the Multi-Spectral dataset: images flipped to compare similarity between palmprints from both hands.}
  \label{fig:fig4}
\end{figure}

We utilized the Gradient-weighted Class Activation Mapping (Grad-CAM) visualization technique~\cite{selvaraju2017grad} for our analysis, as depicted in Fig.~\ref{fig:fig4}. This analysis compared three framework configurations based on the CCNet backbone: Naive, LRPR, and CCPV. The results highlighted notable differences in how each configuration focused on features from left and right palm prints.

The naive and LRPR frameworks demonstrated a significant bias in processing palm print information, struggling to capture features from different orientations accurately. Their feature extraction was suboptimal, leading to poor integration of cross-palm print information. In contrast, the CCPV framework, with its flipping strategy, effectively bridged the modality gap between left and right palm prints. This approach enhanced focus on consistent features across both palm prints, thereby significantly improving cross-modal matching performance.

While the LRPR framework also utilized a flipping strategy, its effectiveness was constrained by the inherent differences between the palm prints. As illustrated in Fig. 4, LRPR managed to capture primary palm print textures to some extent but struggled with the heterogeneous textures between different palms.

In comparison, our proposed CCPV framework excelled in establishing a cross-handed feature space through optimized loss functions. By compressing intra-class features and distinguishing inter-class features, CCPV enhanced the framework’s discriminative ability. This allowed CCPV to preserve attention to prominent and subtle features, capture heterogeneous features across palm prints more effectively, and significantly improve cross-modal matching performance.

\section{Conclusions}

This paper presents a novel CCPV framework, which offers flexibility by allowing the substitution of its recognition backbone with any deep palmprint recognition network. CCPV provides an efficient solution for identity authentication, featuring a key capability: it performs cross-palm verification even if only one palmprint is enrolled. This means it can match palmprints from different hands of the same individual. Additionally, the paper introduces a four-match rule, which elucidates the similarity between reversed palmprint structures and their likelihood of originating from the same hand. Combined with a novel CC loss function, this rule establishes a discriminative feature space that enhances the verification process. Extensive experiments on public datasets validate the effectiveness and robustness of CCPV, demonstrating its superior performance in both open-set and closed-set scenarios.



\bibliographystyle{IEEEbib}
\bibliography{ref}

\end{document}